\documentclass[journal]{IEEEtran}
\usepackage{amsmath}
\usepackage{graphicx}
\usepackage{threeparttable}
\usepackage{multirow}
\usepackage{colortbl}
\usepackage{color}

\usepackage{caption}
\usepackage{subfig}

\usepackage{dblfloatfix}
\usepackage{afterpage}

\usepackage{soul}
\renewcommand{\hl}[1]{#1}

\ifCLASSINFOpdf
\else
\fi
\usepackage[english]{babel}
\usepackage[utf8x]{inputenc}
\usepackage[T1]{fontenc}
\usepackage{amsmath}
\usepackage{amssymb}
\usepackage{booktabs}

\usepackage{amsmath}
\usepackage{graphicx}
\usepackage[colorinlistoftodos]{todonotes}
\usepackage{hyperref}
\newcommand{\subfigure}{\subfloat}
\usepackage{multirow}
\usepackage{amsmath}

\renewcommand{\phi}{\varphi}

\newcommand{\COVAR}[2]{\mathrm{Cov}\left[{#1,#2}\right]}
\newcommand{\Corr}[2]{\mathrm{CorrCoef}\left[{#1,#2}\right]}
\newcommand{\CorrCoef}{\mathrm{Corr}}

\newcommand{\Warp}[2]{\mathrm{warp}\left({#1,#2}\right)}

\newcommand{\Orthogonal}{\mathrm{ortho}}
\newcommand{\Transpose}{\mathrm{T}}
\newcommand{\TV}{\mathrm{TV}}

\newcommand{\Inv}{\mathrm{inv}}

\newcommand{\ModelNameLong}{Volume Tweening Network}
\newcommand{\ModelNameShort}{VTN}
\newcommand{\ModelNameLongShort}{Volume Tweening Network (VTN)}

\newcommand{\emphModelNameLongShort}{\emph{Volume Tweening Network} (VTN)}
\newcommand{\equationref}[1]{Equation~(\ref{#1})}
\newcommand{\figureref}[1]{Figure~\ref{#1}}
\newcommand{\sectionref}[1]{Section~\ref{#1}}
\newcommand{\tableref}[1]{Table~\ref{#1}}

\newcommand{\SpeedUpGpu}{880x}
\newcommand{\SpeedUpCpu}{3.3x}
\newcommand{\xtraAntsAffine}{${}^\star$}

\title{Unsupervised 3D End-to-End Medical Image Registration with {\ModelNameLong}}
\author{Shengyu Zhao\textsuperscript{\textdagger}, Tingfung Lau\textsuperscript{\textdagger}, Ji Luo\textsuperscript{\textdagger}, Eric I-Chao Chang, Yan Xu\textsuperscript{\textasteriskcentered}%
\thanks{This work is supported by the National Science and Technology Major Project of the Ministry of Science and Technology in China under Grant 2017YFC0110903,  National Science and Technology Major Project of the Ministry of Science and Technology in China under Grant 2017YFC0110503, Microsoft Research under the eHealth program, the National Natural Science Foundation in China under Grant 81771910, the Beijing Natural Science Foundation in China under Grant 4152033, the Technology and Innovation Commission of Shenzhen in China under Grant shenfagai2016-627, Beijing Young Talent Project in China, the Fundamental Research Funds for the Central Universities of China under Grant SKLSDE-2017ZX-08 from the State Key Laboratory of Software Development Environment in Beihang University in China, the 111 Project in China under Grant B13003. \emph{Dagger indicates equal contribution. Asterisk indicates the corresponding author.}}%
\thanks{Shengyu Zhao, Tingfung Lau, and Ji Luo are with Institute for Interdisciplinary Information Sciences, Tsinghua University, Beijing 100084, China (email: zsyzzsoft@gmail.com; tingflau@outlook.com; geelaw@outlook.com).}%
\thanks{Yan Xu is with State Key Laboratory of Software Development Environment and Key Laboratory of Biomechanics and Mechanobiology of Ministry of Education and Research Institute of Beihang University in Shenzhen, Beihang University, Beijing 100191, China (email: xuyan04@gmail.com).}%
\thanks{Eric I-Chao Chang and Yan Xu are with Microsoft Research, Beijing 100080, China (email: echang@microsoft.com; xuyan04@gmail.com).}%
\thanks{This work has been submitted to the IEEE for possible publication. Copyright may be transferred without notice, after which this version may no longer be accessible.}%
}

\date{\today}

\begin{document}
\maketitle

\newcommand{\HeroFigure}[1]{
\begin{figure}[#1]
\centering
\subfigure[fixed]
{
	\label{hero-fixed}
	\includegraphics[width=0.6in]{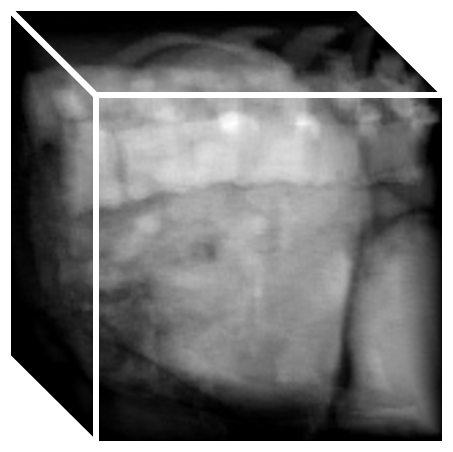}
}
\hspace{0.1in}
\subfigure[moving]
{
	\label{hero-moving}
	\includegraphics[width=0.6in]{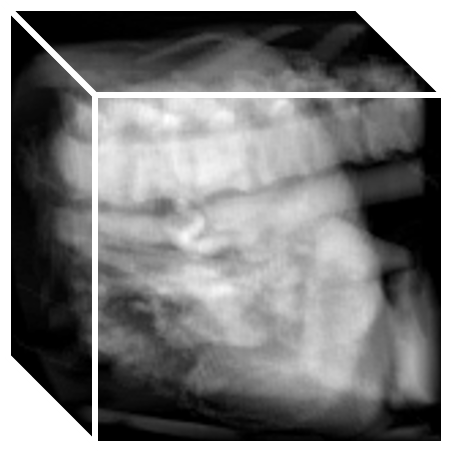}
}
\hspace{0.1in}
\subfigure[deformation]
{
	\label{hero-flow}
	\includegraphics[width=0.6in]{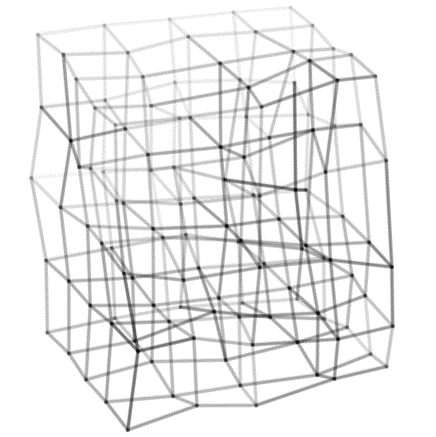}
}
\hspace{0.1in}
\subfigure[warped]
{
	\label{hero-warped}
	\includegraphics[width=0.6in]{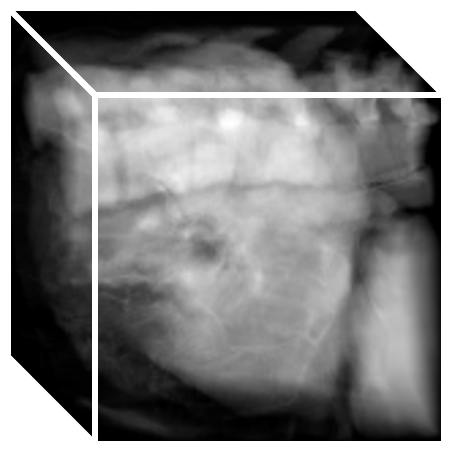}
}
\caption{Illustration of 3D medical image registration. Given a fixed image (a) and a moving image (b), a deformation field (c) indicates the displacement of each voxel in the fixed image to the moving image (represented by a grid skewed according to the field). An image (d) similar to the fixed one can be produced by warping the moving image with the flow. The images are rendered by mapping grayscale to white with transparency (the more intense, the more opaque) and dimetrically projecting the volume.}
\label{hero-example}
\end{figure}
}

\newcommand{\FlowChartFigure}[1]{
\begin{figure*}[#1]
\centering
\includegraphics[width=510pt]{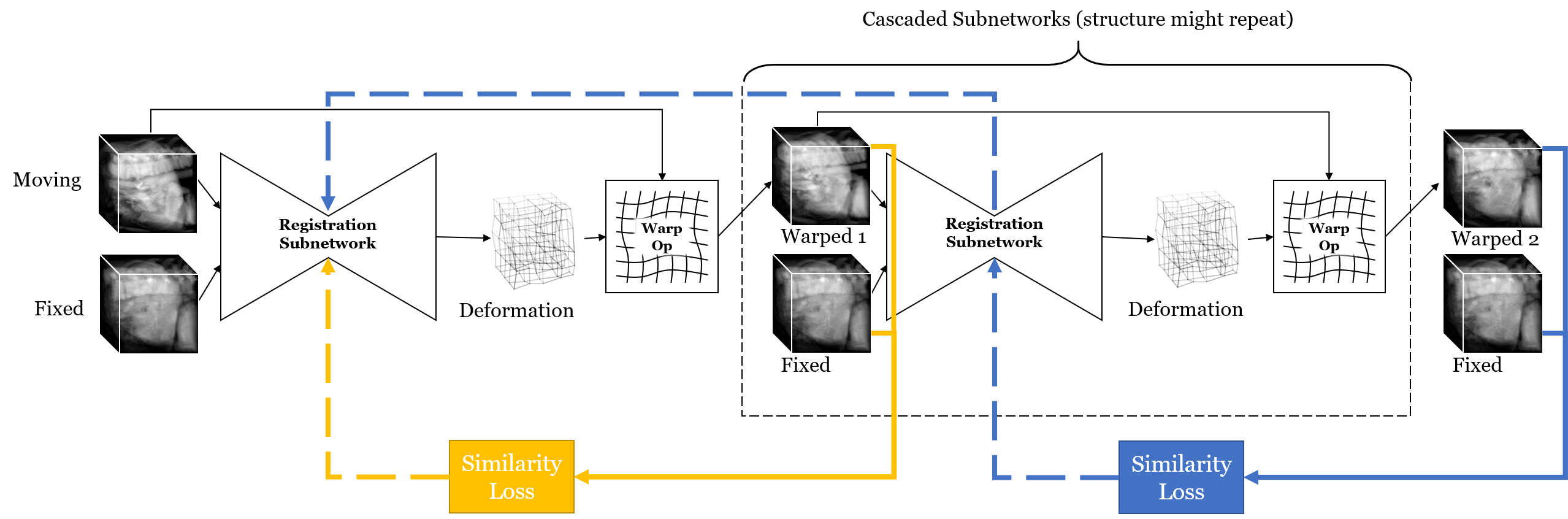}
\caption{Illustration of the overall structure of {\ModelNameLongShort} and how gradients back-propagate. Every registration subnetwork is responsible for finding the deformation field between the fixed image and the current moving image. The moving image is repeatedly warped according to the deformation field and fed into the next level of cascaded subnetworks. The current moving images are compared against the fixed image for a similarity loss function to guide training. There is also regularization loss, but not drawn for the sake of cleanness. The number of cascaded subnetworks may vary; only two are illustrated here. In the figure, yellow (lighter) indicates the similarity loss between the first warped image and the fixed image, and blue (darker) that between the second warped image and the fixed image. Solid bold arrows indicate how the loss is computed, and dashed bold arrows indicate how gradients back-propagate. Note that the second loss will propagate gradients to the first subnetwork as a consequence of the first warped image being a differentiable function of the first subnetwork.}
\label{flowchart}
\label{wholenet}
\end{figure*}
}

\newcommand{\STNFigure}[1]{
\begin{figure}[#1]
\centering
\subfigure[STN structure from \cite{stn}]{\includegraphics[width=3.3in]{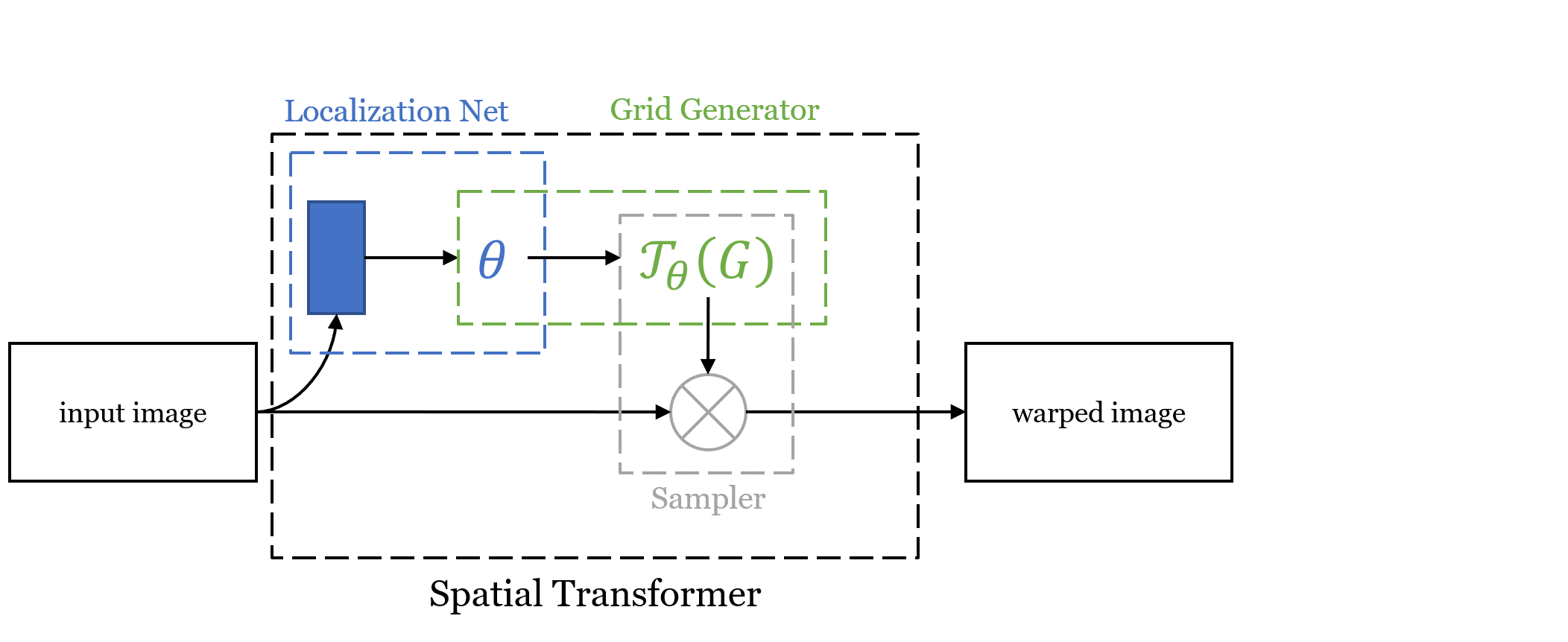}}
\\
\subfigure[fitting our subnetwork into STN parlance]{\includegraphics[width=3.3in]{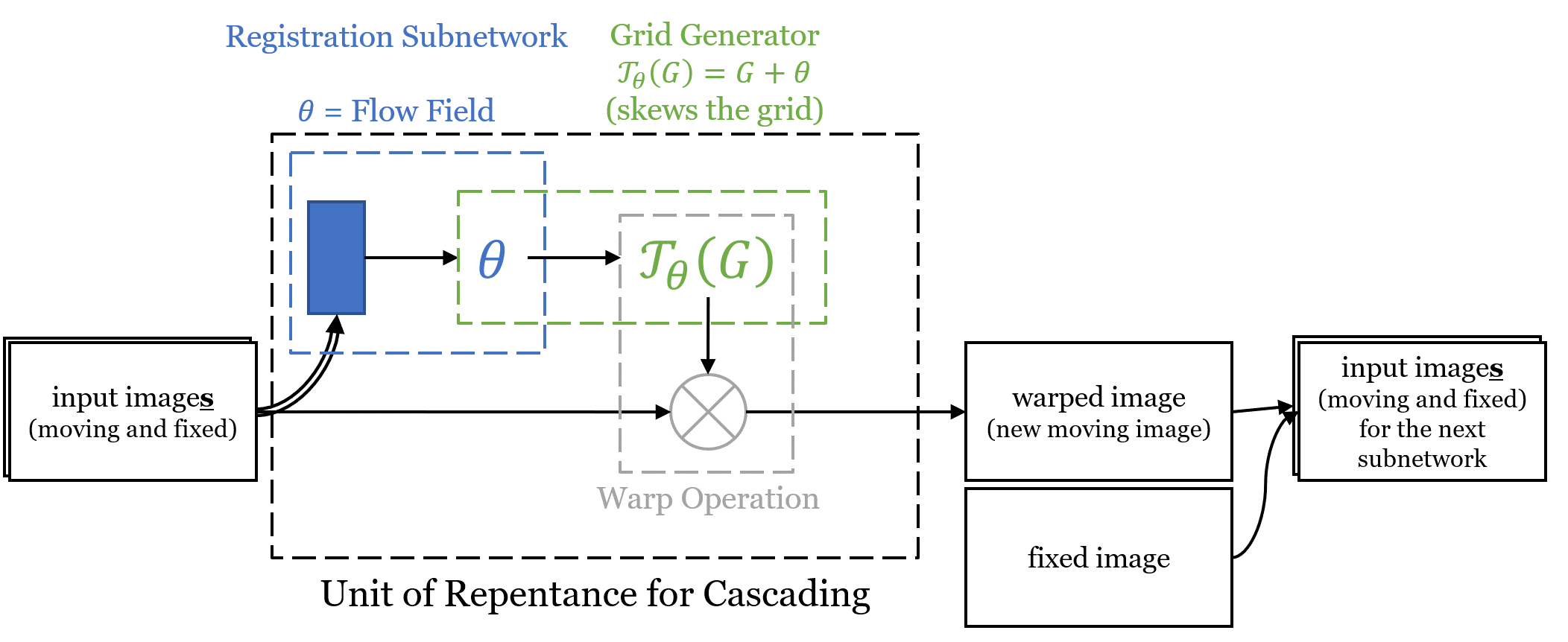}}
\caption{Comparison between STN and a subnetwork in our model. Comparable structures are similarly annotated. However, there is crucial difference between the two, as explained in the article.}
\label{stn-vs-subnet}
\end{figure}
}

\newcommand{\AffineNetFigure}[1]{
\begin{figure}[#1]
\centering
\includegraphics[width=240pt]{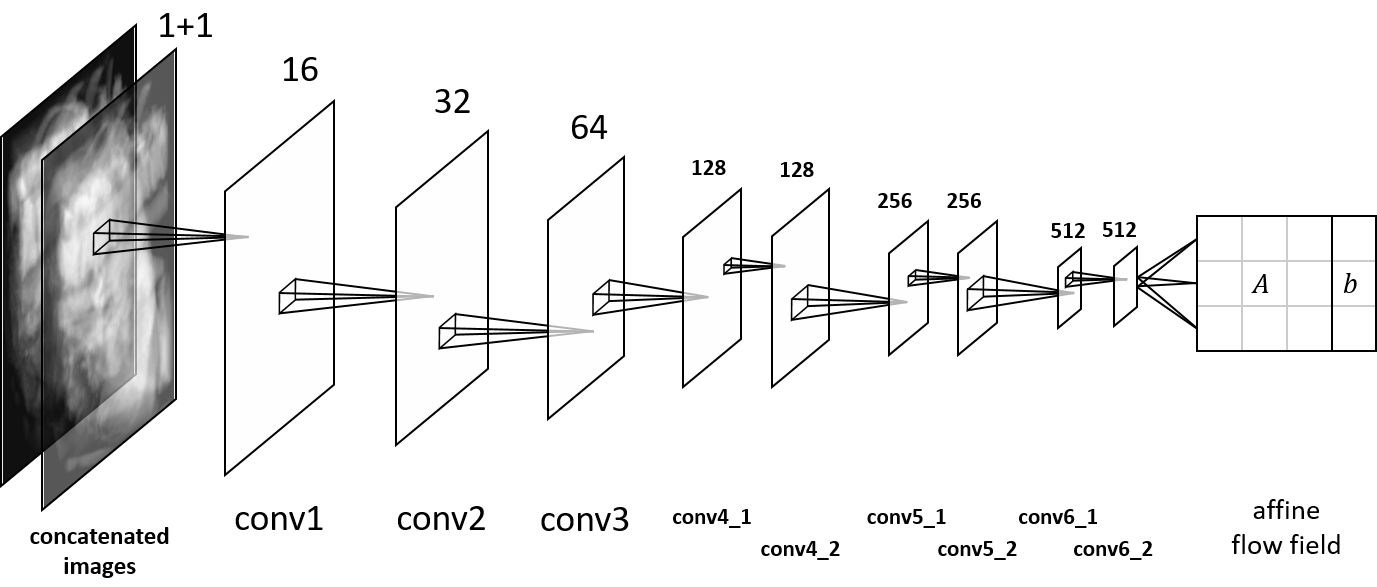}
\caption{Affine registration subnetwork. The number of channels is annotated above the layer. A smaller canvas means lower spatial resolution.}
\label{affinenet}
\end{figure}
}

\newcommand{\DenseNetFigure}[1]{
\begin{figure}[#1]
\centering
\includegraphics[width=240pt]{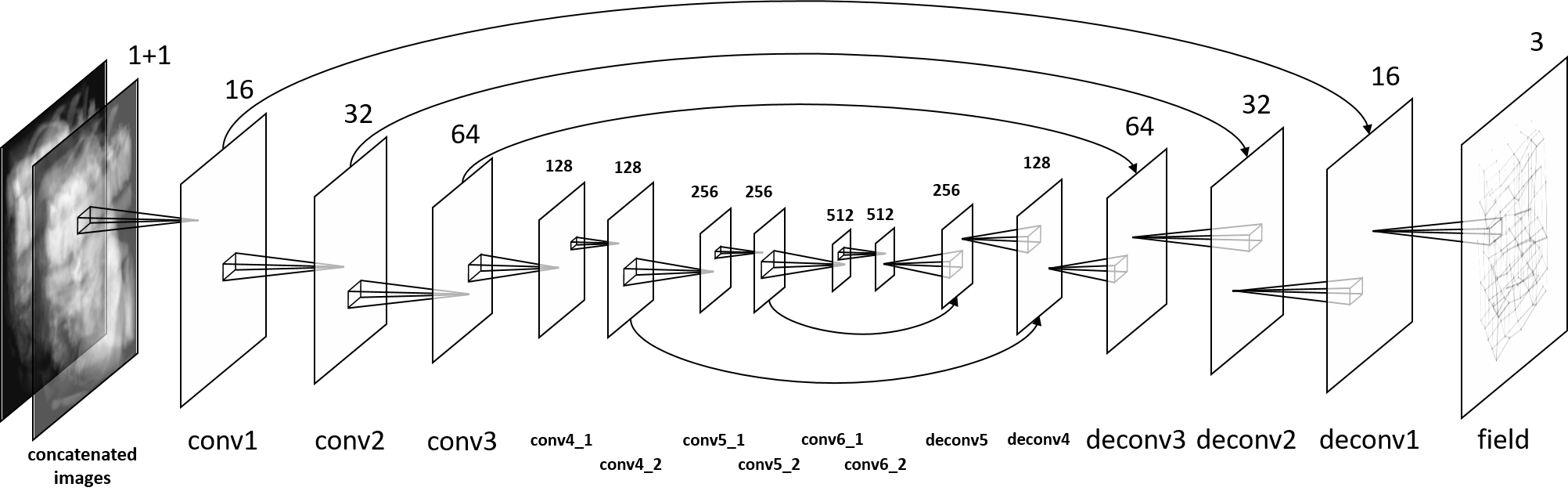}
\caption{Dense deformable registration subnetwork. Number of channels is annotated above the layer. Curved arrows represent skip paths (layers connected by an arrow are concatenated before transposed convolution). Smaller canvas means lower spatial resolution.}
\label{densenet}
\end{figure}
}

\newcommand{\VTNNamingFigure}[1]{
\begin{figure}[#1]
\centering
\subfigure[]{\includegraphics[width=36pt]{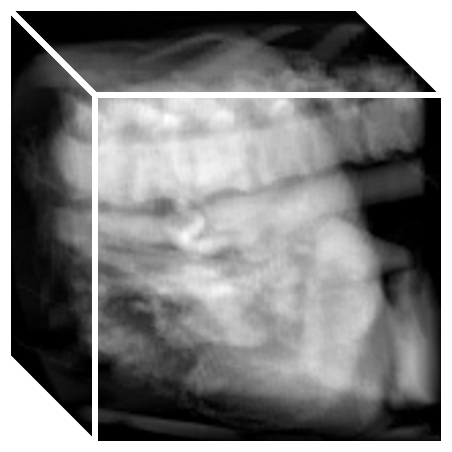}}
~
\subfigure[]{\includegraphics[width=36pt]{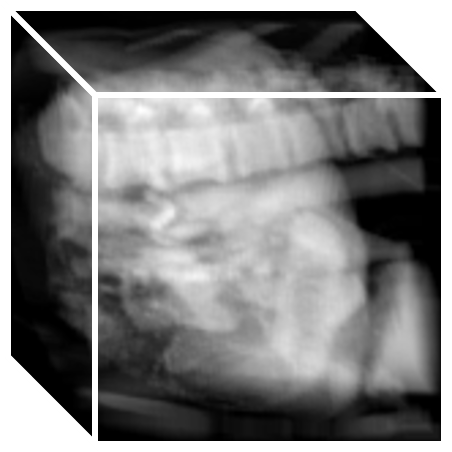}}
~
\subfigure[]{\includegraphics[width=36pt]{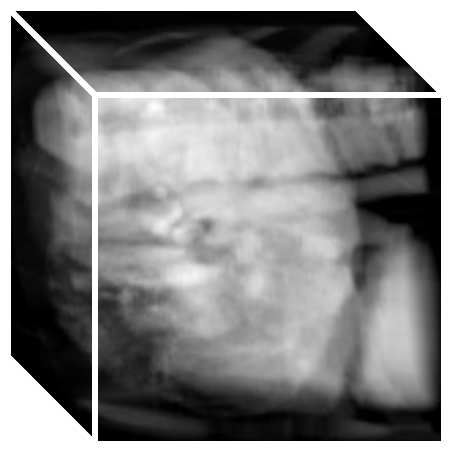}}
~
\subfigure[]{\includegraphics[width=36pt]{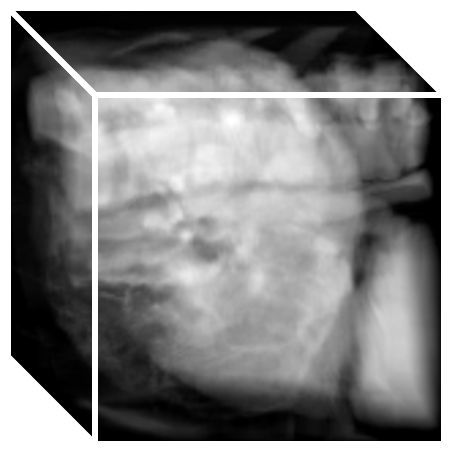}}
~
\subfigure[]{\includegraphics[width=36pt]{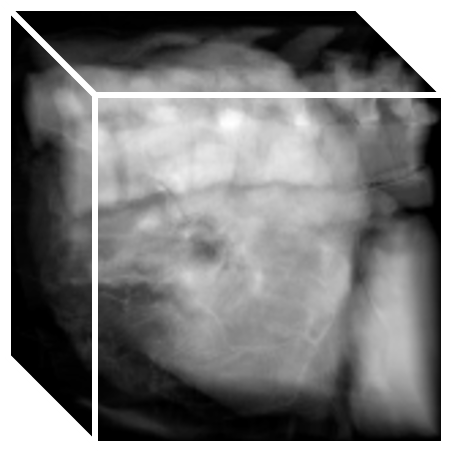}}
~
\subfigure[]{\includegraphics[width=36pt]{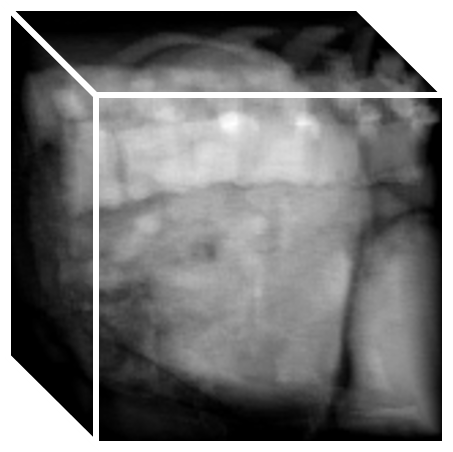}}
\caption{Examples of intermediate warped moving images by a cascaded network consisting of 4 subnetworks. (a) the moving image (a CT liver scan); (b) warped by the affine registration subnetwork; (c/d/e) warped by the first/second/third dense deformable registration subnetwork; (f) the fixed image (another CT liver scan). The images are realistic and are rendered with the same data as those in \figureref{hero-example}.}
\label{vtn-naming}
\end{figure}
}

\newcommand{\LiverMethodsFigure}[1]{
\begin{figure*}[#1]
\centering
\subfigure{\includegraphics[width=50pt]{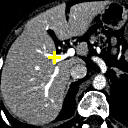}}
~\subfigure{\includegraphics[width=50pt]{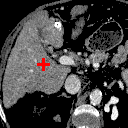}}
~\subfigure{\includegraphics[width=50pt]{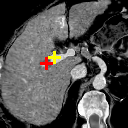}}
~\subfigure{\includegraphics[width=50pt]{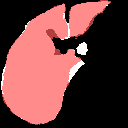}}
~\subfigure{\includegraphics[width=50pt]{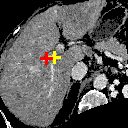}}
~\subfigure{\includegraphics[width=50pt]{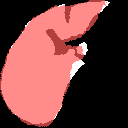}}
~\subfigure{\includegraphics[width=50pt]{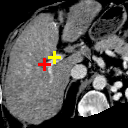}}
~\subfigure{\includegraphics[width=50pt]{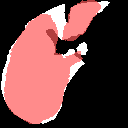}}
~\subfigure{\includegraphics[width=50pt]{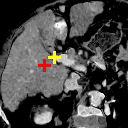}}
~\subfigure{\includegraphics[width=50pt]{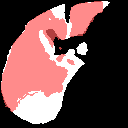}}

\vspace{-7.5pt}
\subfigure{\includegraphics[width=50pt]{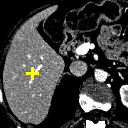}}
~\subfigure{\includegraphics[width=50pt]{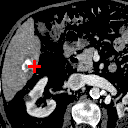}}
~\subfigure{\includegraphics[width=50pt]{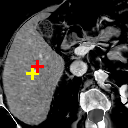}}
~\subfigure{\includegraphics[width=50pt]{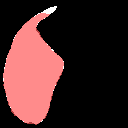}}
~\subfigure{\includegraphics[width=50pt]{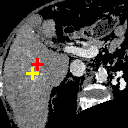}}
~\subfigure{\includegraphics[width=50pt]{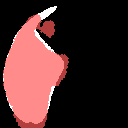}}
~\subfigure{\includegraphics[width=50pt]{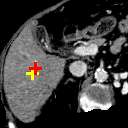}}
~\subfigure{\includegraphics[width=50pt]{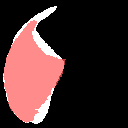}}
~\subfigure{\includegraphics[width=50pt]{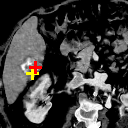}}
~\subfigure{\includegraphics[width=50pt]{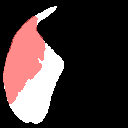}}

\addtocounter{subfigure}{-20}
\vspace{-7.5pt}
\subfigure[]{\includegraphics[width=50pt]{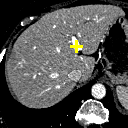}}
~\subfigure[]{\includegraphics[width=50pt]{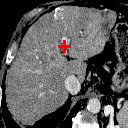}}
~\subfigure[12.49]{\includegraphics[width=50pt]{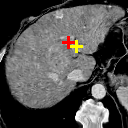}}
~\subfigure[0.9420]{\includegraphics[width=50pt]{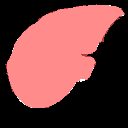}}
~\subfigure[12.44]{\includegraphics[width=50pt]{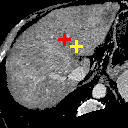}}
~\subfigure[0.8593]{\includegraphics[width=50pt]{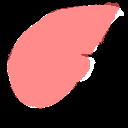}}
~\subfigure[13.28]{\includegraphics[width=50pt]{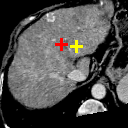}}
~\subfigure[0.8612]{\includegraphics[width=50pt]{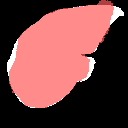}}
~\subfigure[19.20]{\includegraphics[width=50pt]{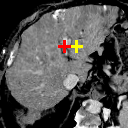}}
~\subfigure[0.7311]{\includegraphics[width=50pt]{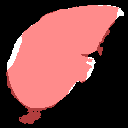}}
\caption{Example comparison among {\ModelNameShort} ADDD + inv (c/d), Elastix (e/f), ANTs (g/h) and VoxelMorph-2 (i/j). (a) sections of the fixed image (a CT liver scan); (b) sections of the moving image (another CT liver scan); (c/e/g/i) sections of the warped images and landmark distances; (d/f/h/j) sections of the warped segmentations (white for the fixed and semi-transparent red for the warped) and segmentation IoUs. Crosses indicate the projection of landmarks (L2, L3 and L4 from top to bottom), yellow (lighter) for one in the fixed image, red (darker) for the corresponding one in the moving/warped images. Best viewed in color.}
\label{liver-algo-fig}
\end{figure*}
}

\newcommand{\LiverCascadeFigure}[1]{
\begin{figure*}[#1]
\centering
\subfigure{\includegraphics[width=41pt]{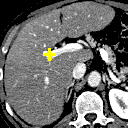}}
~\subfigure{\includegraphics[width=41pt]{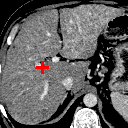}}
~\subfigure{\includegraphics[width=41pt]{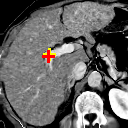}}
~\subfigure{\includegraphics[width=41pt]{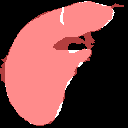}}
~\subfigure{\includegraphics[width=41pt]{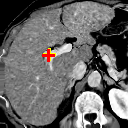}}
~\subfigure{\includegraphics[width=41pt]{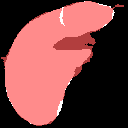}}
~\subfigure{\includegraphics[width=41pt]{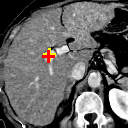}}
~\subfigure{\includegraphics[width=41pt]{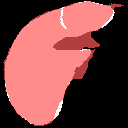}}
~\subfigure{\includegraphics[width=41pt]{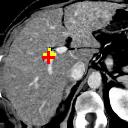}}
~\subfigure{\includegraphics[width=41pt]{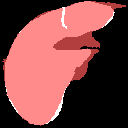}}
~\subfigure{\includegraphics[width=41pt]{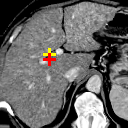}}
~\subfigure{\includegraphics[width=41pt]{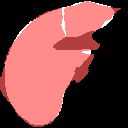}}

\vspace{-7.5pt}
\subfigure{\includegraphics[width=41pt]{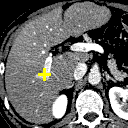}}
~\subfigure{\includegraphics[width=41pt]{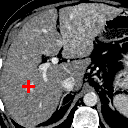}}
~\subfigure{\includegraphics[width=41pt]{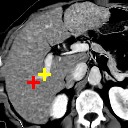}}
~\subfigure{\includegraphics[width=41pt]{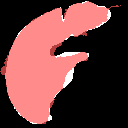}}
~\subfigure{\includegraphics[width=41pt]{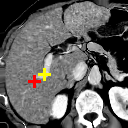}}
~\subfigure{\includegraphics[width=41pt]{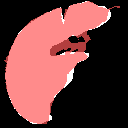}}
~\subfigure{\includegraphics[width=41pt]{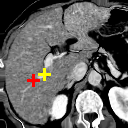}}
~\subfigure{\includegraphics[width=41pt]{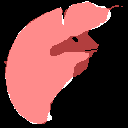}}
~\subfigure{\includegraphics[width=41pt]{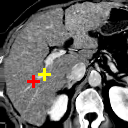}}
~\subfigure{\includegraphics[width=41pt]{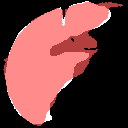}}
~\subfigure{\includegraphics[width=41pt]{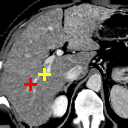}}
~\subfigure{\includegraphics[width=41pt]{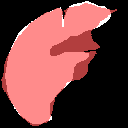}}

\addtocounter{subfigure}{-24}
\vspace{-7.5pt}
\subfigure[]{\includegraphics[width=41pt]{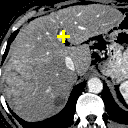}}
~\subfigure[]{\includegraphics[width=41pt]{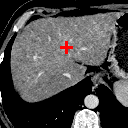}}
~\subfigure[8.80]{\includegraphics[width=41pt]{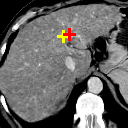}}
~\subfigure[0.9045]{\includegraphics[width=41pt]{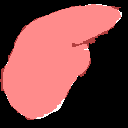}}
~\subfigure[7.80]{\includegraphics[width=41pt]{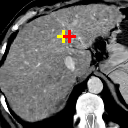}}
~\subfigure[0.8999]{\includegraphics[width=41pt]{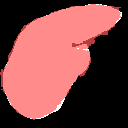}}
~\subfigure[9.00]{\includegraphics[width=41pt]{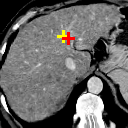}}
~\subfigure[0.8810]{\includegraphics[width=41pt]{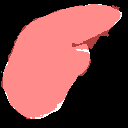}}
~\subfigure[9.21]{\includegraphics[width=41pt]{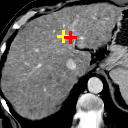}}
~\subfigure[0.8394]{\includegraphics[width=41pt]{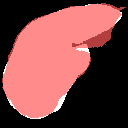}}
~\subfigure[12.79]{\includegraphics[width=41pt]{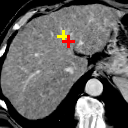}}
~\subfigure[0.8265]{\includegraphics[width=41pt]{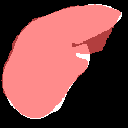}}
\caption{\hl{Example comparison among ADDD + inv (c/d), ADDD (e/f), ADD (g/h), AD (i/j) and D (k/l) networks.}}
\label{liver-cascade-fig}
\end{figure*}
}

\newcommand{\LiverCascadeProgressFigure}[1]{
\begin{figure*}[#1]
\centering
\subfigure{\includegraphics[width=50pt]{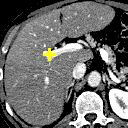}}
~\subfigure{\includegraphics[width=50pt]{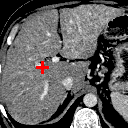}}
~\subfigure{\includegraphics[width=50pt]{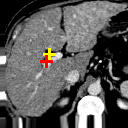}}
~\subfigure{\includegraphics[width=50pt]{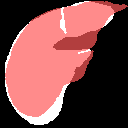}}
~\subfigure{\includegraphics[width=50pt]{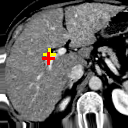}}
~\subfigure{\includegraphics[width=50pt]{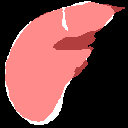}}
~\subfigure{\includegraphics[width=50pt]{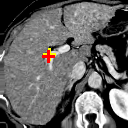}}
~\subfigure{\includegraphics[width=50pt]{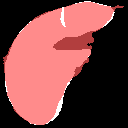}}
~\subfigure{\includegraphics[width=50pt]{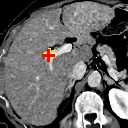}}
~\subfigure{\includegraphics[width=50pt]{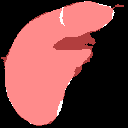}}

\vspace{-7.5pt}
\subfigure{\includegraphics[width=50pt]{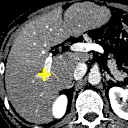}}
~\subfigure{\includegraphics[width=50pt]{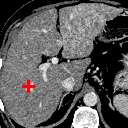}}
~\subfigure{\includegraphics[width=50pt]{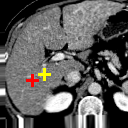}}
~\subfigure{\includegraphics[width=50pt]{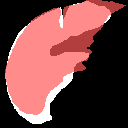}}
~\subfigure{\includegraphics[width=50pt]{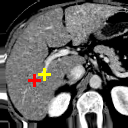}}
~\subfigure{\includegraphics[width=50pt]{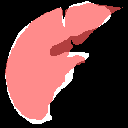}}
~\subfigure{\includegraphics[width=50pt]{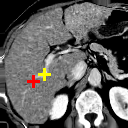}}
~\subfigure{\includegraphics[width=50pt]{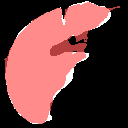}}
~\subfigure{\includegraphics[width=50pt]{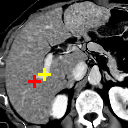}}
~\subfigure{\includegraphics[width=50pt]{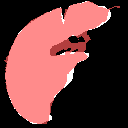}}

\addtocounter{subfigure}{-20}
\vspace{-7.5pt}
\subfigure[]{\includegraphics[width=50pt]{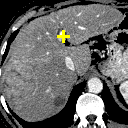}}
~\subfigure[]{\includegraphics[width=50pt]{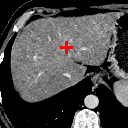}}
~\subfigure[11.81]{\includegraphics[width=50pt]{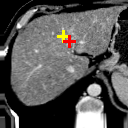}}
~\subfigure[0.6295]{\includegraphics[width=50pt]{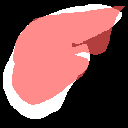}}
~\subfigure[9.19]{\includegraphics[width=50pt]{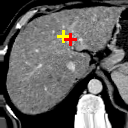}}
~\subfigure[0.7613]{\includegraphics[width=50pt]{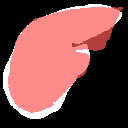}}
~\subfigure[8.56]{\includegraphics[width=50pt]{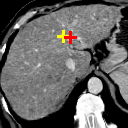}}
~\subfigure[0.8681]{\includegraphics[width=50pt]{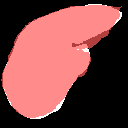}}
~\subfigure[7.80]{\includegraphics[width=50pt]{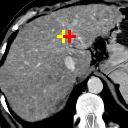}}
~\subfigure[0.8999]{\includegraphics[width=50pt]{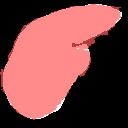}}
\caption{Example intermediate warped moving images by ADDD network. (c/d) warped by the affine subnetwork; (e/f/g/h/i/j) warped by the first/second/third dense deformable subnetwork. Columning and coloring are the same as those in \figureref{liver-algo-fig}, except that the fixed image and the moving image are another pair of CT liver scans. Best viewed in color.}
\label{liver-cascade-progress-fig}
\end{figure*}
}

\newcommand{\BrainMethodsFigure}[1]{
\begin{figure*}[#1]
\centering
\subfigure{\includegraphics[width=41pt]{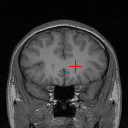}}
~\subfigure{\includegraphics[width=41pt]{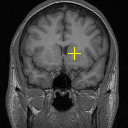}}
~\subfigure{\includegraphics[width=41pt]{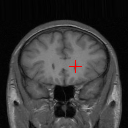}}
~\subfigure{\includegraphics[width=41pt]{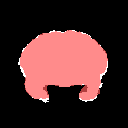}}
~\subfigure{\includegraphics[width=41pt]{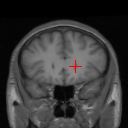}}
~\subfigure{\includegraphics[width=41pt]{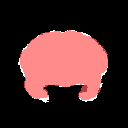}}
~\subfigure{\includegraphics[width=41pt]{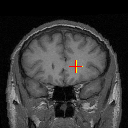}}
~\subfigure{\includegraphics[width=41pt]{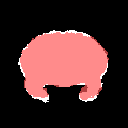}}
~\subfigure{\includegraphics[width=41pt]{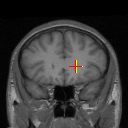}}
~\subfigure{\includegraphics[width=41pt]{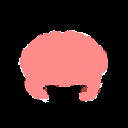}}
~\subfigure{\includegraphics[width=41pt]{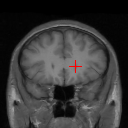}}
~\subfigure{\includegraphics[width=41pt]{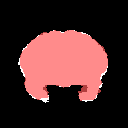}}

\vspace{-7.5pt}
\subfigure{\includegraphics[width=41pt]{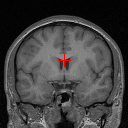}}
~\subfigure{\includegraphics[width=41pt]{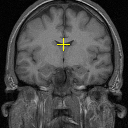}}
~\subfigure{\includegraphics[width=41pt]{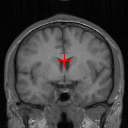}}
~\subfigure{\includegraphics[width=41pt]{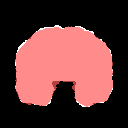}}
~\subfigure{\includegraphics[width=41pt]{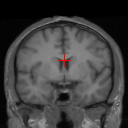}}
~\subfigure{\includegraphics[width=41pt]{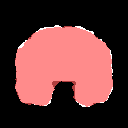}}
~\subfigure{\includegraphics[width=41pt]{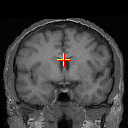}}
~\subfigure{\includegraphics[width=41pt]{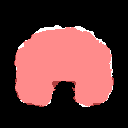}}
~\subfigure{\includegraphics[width=41pt]{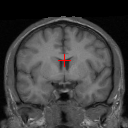}}
~\subfigure{\includegraphics[width=41pt]{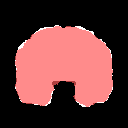}}
~\subfigure{\includegraphics[width=41pt]{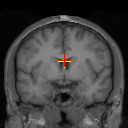}}
~\subfigure{\includegraphics[width=41pt]{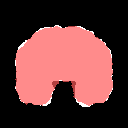}}

\addtocounter{subfigure}{-24}
\vspace{-7.5pt}
\subfigure[]{\includegraphics[width=41pt]{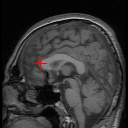}}
~\subfigure[]{\includegraphics[width=41pt]{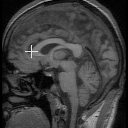}}
~\subfigure[2.06]{\includegraphics[width=41pt]{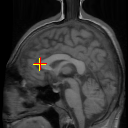}}
~\subfigure[0.9414]{\includegraphics[width=41pt]{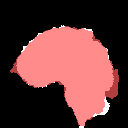}}
~\subfigure[2.10]{\includegraphics[width=41pt]{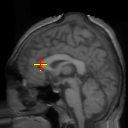}}
~\subfigure[0.9408]{\includegraphics[width=41pt]{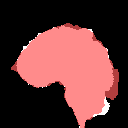}}
~\subfigure[2.46]{\includegraphics[width=41pt]{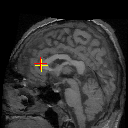}}
~\subfigure[0.9211]{\includegraphics[width=41pt]{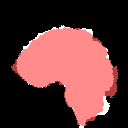}}
~\subfigure[2.18]{\includegraphics[width=41pt]{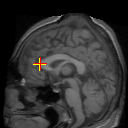}}
~\subfigure[0.9439]{\includegraphics[width=41pt]{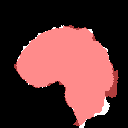}}
~\subfigure[2.27]{\includegraphics[width=41pt]{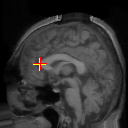}}
~\subfigure[0.9345]{\includegraphics[width=41pt]{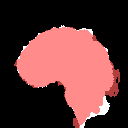}}
\caption{Example comparison among {\ModelNameShort} ADD (c/d), {\ModelNameShort} DD\xtraAntsAffine (e/f), Elastix (g/h), ANTs (i/j) and VoxelMorph-2\xtraAntsAffine (k/l). The input images to methods with ``\xtraAntsAffine'' are affinely aligned to a fixed atlas by ANTs and their warped images are transformed backwards according to the affine transformation aligning the fixed image and the atlas for sensible comparison. Columning and coloring are the same as those in \figureref{liver-algo-fig}, except that the fixed image and the moving image are a pair of MR brain scans and that the landmarks are L7, L12 and L15. Best viewed in color.}
\label{brain-algo-fig}
\end{figure*}
}

\newcommand{\BrainSupFigure}[1]{
\begin{figure*}[#1]
\centering
\subfigure{\includegraphics[width=62pt]{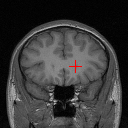}}
~\subfigure{\includegraphics[width=62pt]{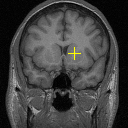}}
~\subfigure{\includegraphics[width=62pt]{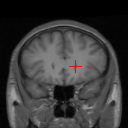}}
~\subfigure{\includegraphics[width=62pt]{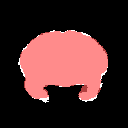}}
~\subfigure{\includegraphics[width=62pt]{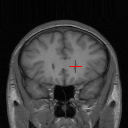}}
~\subfigure{\includegraphics[width=62pt]{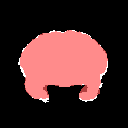}}
~\subfigure{\includegraphics[width=62pt]{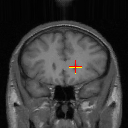}}
~\subfigure{\includegraphics[width=62pt]{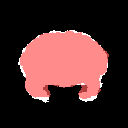}}

\vspace{-7.5pt}
\subfigure{\includegraphics[width=62pt]{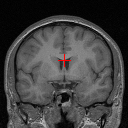}}
~\subfigure{\includegraphics[width=62pt]{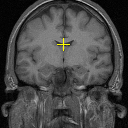}}
~\subfigure{\includegraphics[width=62pt]{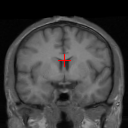}}
~\subfigure{\includegraphics[width=62pt]{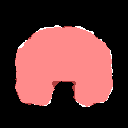}}
~\subfigure{\includegraphics[width=62pt]{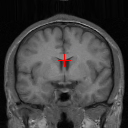}}
~\subfigure{\includegraphics[width=62pt]{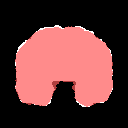}}
~\subfigure{\includegraphics[width=62pt]{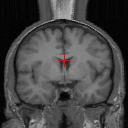}}
~\subfigure{\includegraphics[width=62pt]{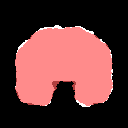}}

\addtocounter{subfigure}{-16}
\vspace{-7.5pt}
\subfigure[]{\includegraphics[width=62pt]{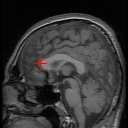}}
~\subfigure[]{\includegraphics[width=62pt]{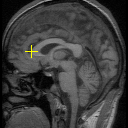}}
~\subfigure[2.10]{\includegraphics[width=62pt]{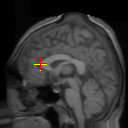}}
~\subfigure[0.9408]{\includegraphics[width=62pt]{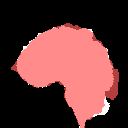}}
~\subfigure[2.06]{\includegraphics[width=62pt]{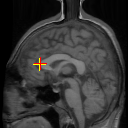}}
~\subfigure[0.9414]{\includegraphics[width=62pt]{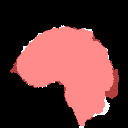}}
~\subfigure[2.37]{\includegraphics[width=62pt]{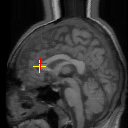}}
~\subfigure[0.9222]{\includegraphics[width=62pt]{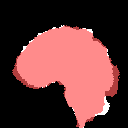}}
\caption{Example comparison among unsupervised and supervised {\ModelNameShort}. (c/d) DD\xtraAntsAffine; (e/f) ADD; (g/h) ADD (supervised). Rendering, columning and coloring are the same as those in \figureref{brain-algo-fig}. Best viewed in color.}
\label{brain-sup-fig}
\end{figure*}
}

\newcommand{\BrainCascadeFigure}[1]{
\begin{figure*}[#1]
\centering
\subfigure{\includegraphics[width=50pt]{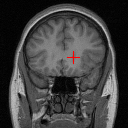}}
~\subfigure{\includegraphics[width=50pt]{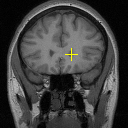}}
~\subfigure{\includegraphics[width=50pt]{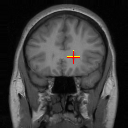}}
~\subfigure{\includegraphics[width=50pt]{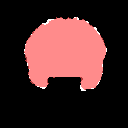}}
~\subfigure{\includegraphics[width=50pt]{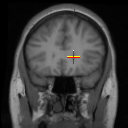}}
~\subfigure{\includegraphics[width=50pt]{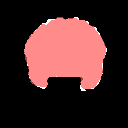}}
~\subfigure{\includegraphics[width=50pt]{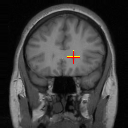}}
~\subfigure{\includegraphics[width=50pt]{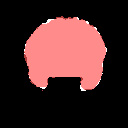}}
~\subfigure{\includegraphics[width=50pt]{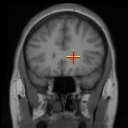}}
~\subfigure{\includegraphics[width=50pt]{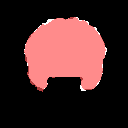}}

\vspace{-7.5pt}
\subfigure{\includegraphics[width=50pt]{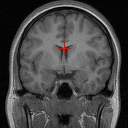}}
~\subfigure{\includegraphics[width=50pt]{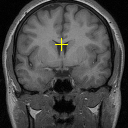}}
~\subfigure{\includegraphics[width=50pt]{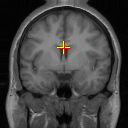}}
~\subfigure{\includegraphics[width=50pt]{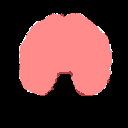}}
~\subfigure{\includegraphics[width=50pt]{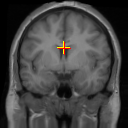}}
~\subfigure{\includegraphics[width=50pt]{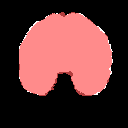}}
~\subfigure{\includegraphics[width=50pt]{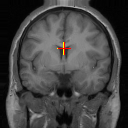}}
~\subfigure{\includegraphics[width=50pt]{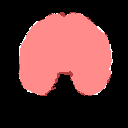}}
~\subfigure{\includegraphics[width=50pt]{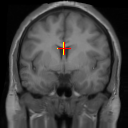}}
~\subfigure{\includegraphics[width=50pt]{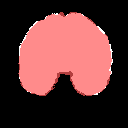}}

\addtocounter{subfigure}{-20}
\vspace{-7.5pt}
\subfigure[]{\includegraphics[width=50pt]{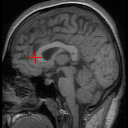}}
~\subfigure[]{\includegraphics[width=50pt]{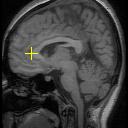}}
~\subfigure[2.24]{\includegraphics[width=50pt]{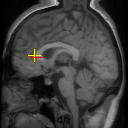}}
~\subfigure[0.9446]{\includegraphics[width=50pt]{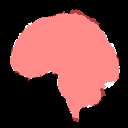}}
~\subfigure[2.48]{\includegraphics[width=50pt]{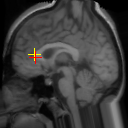}}
~\subfigure[0.9193]{\includegraphics[width=50pt]{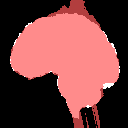}}
~\subfigure[2.47]{\includegraphics[width=50pt]{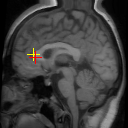}}
~\subfigure[0.9367]{\includegraphics[width=50pt]{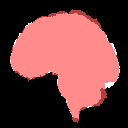}}
~\subfigure[2.38]{\includegraphics[width=50pt]{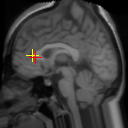}}
~\subfigure[0.9087]{\includegraphics[width=50pt]{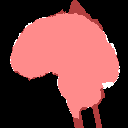}}
\caption{Example comparison among {\ModelNameShort} with and without integrated affine registration. (c/d) ADD; (e/f) {DD\xtraAntsAffine}; (g/h) AD; (i/j) {D\xtraAntsAffine}. Rendering, columning and coloring are the same as those in \figureref{brain-algo-fig}, except that the fixed image and the moving image are another pair of MR brain scans. Best viewed in color.}
\label{brain-cascade-fig}
\end{figure*}
}

\newcommand{\InvLossFigure}[1]{
\begin{figure}[#1]
\centering
\includegraphics[width=1.5in]{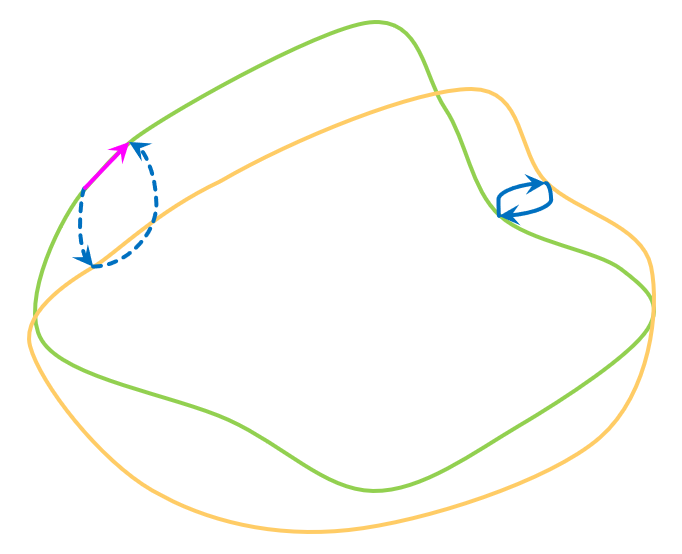}
\caption{Illustration of how invertibility loss enforces round-trip registration. Green (darker) and yellow (lighter) curves represent two images. Curved arrows: flow fields. Solid curved arrows: flow inversion. Dashed arrows: failure of flow inverseion. Straight arrow (magenta, very dark): example vector in the composed flow green $\to$ yellow $\to$ green, which is non-zero because the voxel fails to trip back. Best viewed in color.}
\label{invloss-fig}
\end{figure}
}

\newcommand{\LossRatioTable}[1]{
\begin{table}[#1]
\centering
\begin{tabular}{@{}ccc@{}}
\toprule
Subnetwork & Loss & Relative Ratio
\\ \midrule
Affine & Similarity & 1 \\
       & Determinant & 0.1 \\
       & Orthogonality & 0.1 \\ \midrule
Dense 1 & Similarity & 0 \\
        & Total variation & 1 \\ \midrule
Dense 2 & Similarity & 0.05 \\
        & Total variation & 1 \\ \midrule
Dense 3 & Similarity & 1 \\
        & Total variation & 1 \\
\bottomrule
\end{tabular}
\caption{Ratio of loss functions.}
\label{loss-ratio-table}
\end{table}
}

\newcommand{\LiverMethodsTable}[1]{
\begin{table}[#1]
\centering
\begin{tabular}{@{}ccccc@{}}
\toprule
Method & Seg. IoU & Lm. Dist. & Time & Time (w/o GPU)
\\ \midrule
ANTs \cite{avants2009advanced} & 0.8124 & 11.93 & N/A & 748 s \\
Elastix \cite{elastix} & 0.8365 & 12.36 & N/A  & 115 s \\
VoxelMorph-2 \cite{voxelmorph} & 0.6796 & 18.10 & 0.20 s & 17 s \\
{\ModelNameShort} ADDD & 0.8868 & 12.04 & 0.13 s & 26 s \\
{\ModelNameShort} ADDD + inv & \textbf{0.8882} & \textbf{11.42} & 0.13 s & 26 s \\ \hline
supervised & 0.7680 & 13.38 & 0.13 s & 26 s \\
\bottomrule
\end{tabular}
\caption{Comparison among traditional methods, VoxelMorph and our {\ModelNameShort} (liver).}
\label{liver-algo-table}
\end{table}
}

\newcommand{\LiverJacTable}[1]{
\begin{table}[#1]
\centering
\begin{tabular}{@{}ccc@{}}
\toprule
Method & Std. Jacobian & Folding (\%)
\\ \midrule
ANTs \cite{avants2009advanced} & 0.241 (0.030) & 0.194 (0.145) \\
Elastix \cite{elastix} & 0.264 (0.072) & 0.001 (0.006) \\
{\ModelNameShort} ADDD & 0.787 (0.171) & 0.365 (0.268) \\
{\ModelNameShort} ADDD + inv & 0.770 (0.169) & 0.288 (0.240) \\
\bottomrule
\end{tabular}
\caption{\hl{Standard deviation of Jacobian determinant and fraction of folding on the liver dataset. Areas with negative Jacobian determinant are considered folding. Standard deviations across instances are in parentheses.}}
\label{liver-jac-table}
\end{table}
}

\newcommand{\LiverMoreDataTable}[1]{
\begin{table}[#1]
\centering
\begin{tabular}{@{}lcc@{}}
\toprule
Training Dataset for {\ModelNameShort} ADDD & Seg. IoU & Lm. Dist.
\\ \midrule
LITS & 0.8712 & 13.11 \\
\hl{BFH} & \hl{0.8815} & \hl{12.78} \\
LITS + BFH & \textbf{0.8868} & \textbf{12.04} \\
\bottomrule
\end{tabular}
\caption{Comparison of performance of {\ModelNameShort} ADDD with different amount of unlabeled training data (liver).}
\label{liver-moredata-table}
\end{table}
}

\newcommand{\LiverCascadeTable}[1]{
\begin{table}[#1]
\centering
\begin{tabular}{@{}rcccc@{}}
\toprule
Network & Seg. IoU & Lm. Dist. & Time & Time (w/o GPU)
\\ \midrule
D    & 0.8119 & 14.44 & 0.08 s & 10 s \\
DD   & 0.8556 & 12.97 & 0.10 s & 20 s \\
DDD  & 0.8709 & 12.49 & 0.12 s & 28 s \\ \midrule
AD   & 0.8323 & 13.20 & 0.09 s & 9 s \\
ADD  & 0.8703 & 12.28 & 0.11 s & 19 s \\
ADDD & \textbf{0.8868} & \textbf{12.04} & 0.13 s & 26 s \\
\bottomrule
\end{tabular}
\caption{Comparison of performance with different number of cascaded subnetworks (liver).}
\label{liver-cascade-table}
\end{table}
}

\newcommand{\BrainMethodsTable}[1]{
\begin{table}[#1]
\centering
\begin{tabular}{@{}ccccc@{}}
\toprule
Method & Seg. IoU & Lm. Dist. & Time & Time (w/o GPU) \\ \midrule
ANTs \cite{avants2009advanced} & \textbf{0.9387} & 2.85 & N/A & 764 s \\
Elastix \cite{elastix} & 0.9180 & 3.23 & N/A & 121 s \\
{VoxelMorph-2\xtraAntsAffine} \cite{voxelmorph} & 0.9268 & 2.84 & 0.19 s & 14 s \\
{\ModelNameShort} {DD\xtraAntsAffine} & 0.9305\xtraAntsAffine & 2.83\xtraAntsAffine & 0.09 s & 19 s \\
{\ModelNameShort} ADD & 0.9270 & \textbf{2.62} & 0.10 s & 17 s \\
{\ModelNameShort} ADD + inv & 0.9278 & 2.64 & 0.10 s & 17 s \\ \hline
supervised & 0.9060 & 2.94 & 0.10 s & 19 s \\
\bottomrule
\end{tabular}
\caption{Comparison among traditional methods, VoxelMorph \cite{voxelmorph} and our algorithm on brain datasets. Bold: best among all methods. Star: best among methods with ANTs affine pre-alignment.\\
Methods with ``{\xtraAntsAffine}'' use ANTs for affine pre-alignment. The preprocessing time (about 74 seconds) is not included in the table.}
\label{brain-algo-table}
\end{table}
}

\newcommand{\BrainJacTable}[1]{
\begin{table}[#1]
\centering
\begin{tabular}{@{}ccc@{}}
\toprule
Method & Std. Jacobian & Folding (\%)
\\ \midrule
ANTs \cite{avants2009advanced} & 0.230 (0.037) & 0.294 (0.228) \\
Elastix \cite{elastix} & 0.171 (0.034) & 0.000 (0.000) \\
{\ModelNameShort} ADD & 0.339 (0.126) & 0.018 (0.043) \\
{\ModelNameShort} ADD + inv & 0.332 (0.122) & 0.017 (0.055) \\
\bottomrule
\end{tabular}
\caption{\hl{Standard deviation of Jacobian determinant and fraction of folding on the brain dataset. Areas with negative Jacobian determinant are considered folding. Standard deviations across instances are in parentheses.}}
\label{brain-jac-table}
\end{table}
}

\newcommand{\BrainMoreDataTable}[1]{
\begin{table}[#1]
\centering
\begin{tabular}{@{}lcc@{}}
\toprule
Training Dataset for {\ModelNameShort} DD & Seg. IoU & Lm. Dist.
\\ \midrule
ADNI + ABIDE-1 & 0.9299 & 2.90 \\
ADNI + ABIDE-1 + ABIDE-2 & \textbf{0.9312} & 2.86 \\
ADNI + ABIDE-1 + ABIDE-2 + ADHD & 0.9305 & \textbf{2.83} \\
\bottomrule
\end{tabular}
\caption{Comparison of performance of {\ModelNameShort} DD (with ANTs affine alignment) with different amount of unlabeled training data (brain).}
\label{brain-moredata-table}
\end{table}
}

\newcommand{\BrainCascadeTable}[1]{
\begin{table}[#1]
\centering
\begin{tabular}{@{}rcccc@{}}
\toprule
Method & Seg. IoU & Lm. Dist. & Time & Time (w/o GPU)
\\ \midrule
{D\xtraAntsAffine}  & 0.9241 & 2.91 & 0.07 s & 10 s \\
{DD\xtraAntsAffine} & 0.9305 & 2.83\xtraAntsAffine & 0.09 s & 19 s \\
{DDD\xtraAntsAffine} & \textbf{0.9320}\xtraAntsAffine & 2.85 & 0.10 s & 28 s \\ \midrule
AD   & 0.9214 & 2.73 & 0.08 s & 9 s \\
ADD  & 0.9270 & \textbf{2.62} & 0.10 s & 17 s \\
ADDD & 0.9286 & 2.63 & 0.11 s & 26 s \\
\bottomrule
\end{tabular}
\caption{Comparison of performance with different number of cascaded subnetworks (brain), and comparison between using ANTs as affine alignment and end-to-end network (integrated affine registration subnetwork). The first group (with ``{\xtraAntsAffine}'') uses ANTs to affinely align input images, whereas the second group does not.  Bold: best among all methods. Star: best among methods with ANTs affine pre-alignment.\\
Methods with ``{\xtraAntsAffine}'' use ANTs for affine pre-alignment. The preprocessing time (about 74 seconds) is not included in the table.}
\label{brain-cascade-table}
\end{table}
}

\begin{abstract}
3D medical image registration is of great clinical importance. However, supervised learning methods require a large amount of accurately annotated corresponding control points (or morphing), which are very difficult to obtain. Unsupervised learning methods ease the burden of manual annotation by exploiting unlabeled data without supervision.
In this paper, we propose a new unsupervised learning method using convolutional neural networks under an end-to-end framework, {\emphModelNameLongShort}, for 3D medical image registration. \hl{We propose three innovative technical components:
(1)~An end-to-end cascading scheme that resolves large displacement;
(2)~An efficient integration of affine registration network; and
(3)~An additional invertibility loss that encourages backward consistency.}
Experiments demonstrate that our algorithm is {\SpeedUpGpu} faster (or {\SpeedUpCpu} faster without GPU acceleration) than traditional optimization-based methods and achieves state-of-the-art performance in medical image registration.
\end{abstract}



\section{Introduction}

\hl{Image registration is the process of finding the non-linear spatial correspondence between the input images} (see \figureref{hero-example}). It has a wide range of applications in medical image processing, such as aligning images of one subject taken at different times. Another example is to match an image of one subject to some predefined coordinate system, such as an anatomical atlas \cite{oliveira2014medical}.

\hl{Over recent decades, many traditional algorithms have been developed and studied to register medical images }\cite{Hermosillo2002,mi-reg-freeform,avants2009advanced,elastix}\hl{. Most of them define a hypothesis space of possible transforms, often described by a set of parameters, and a metric of the quality of such a transform, and then find the optimum by iteratively updating the parameters. Traditional methods have achieved good performance on several datasets, and are state-of-the-art, but their registration speed is barely practical for clinical applications. These methods do not exploit the patterns that exist in the registration task. In contrast, learning-based methods are generally faster. Computationally, they do not need to iterate over and over before producing the result. Conceptually, they learn those patterns represented by the parameters of the underlying learning model. The trained model is an efficient replacement of the optimization procedure.}

However, there has not been an effective approach to generate ground-truth fields for medical images. Widely used supervised methods require accurately labeled ground truth with a vast number of instances. The quality of these labels directly affects the result of supervision, which entails much effort in traditional tasks such as classification and segmentation. \hl{But flow fields are dense and ambiguous quantities that are almost impossible to be labeled manually}, and moreover, automatically generated dataset (e.g., the Flying Chairs dataset \cite{dosovitskiy2015flownet}) which deviates from the realistic demands is not appropriate. Consequently, supervised methods are hardly applicable. In contrast, unlabeled medical images are universally available, and sufficient to advance the state of the art through our unsupervised framework shown in this paper.

\HeroFigure{!t}


\FlowChartFigure{!t}

Optical flow estimation is a closely related problem that aims to identify the correspondence of pixels in two images of the same scene taken from different perspectives. FlowNet \cite{dosovitskiy2015flownet} and its successor FlowNet 2.0 \cite{IMKDB17} are CNNs that predict optical flow from input images using fully convolutional networks (FCN \cite{long2015fully}), which are capable of regressing pixel-level correspondence. FlowNet is trained on Flying Chairs, a synthetic dataset that consists of images generated using computer graphics algorithms and the ground-truth flows.

Spatial Transformer Networks (STN) \cite{stn} is a component in neural networks that spatially transforms feature maps to ease back-end tasks. It learns a localization net to produce an appropriate transformation to ``straighten'' the input image. The localization net is learnt without supervision, though back-end task might be supervised. Given a sampling grid and a transformation, STN applies the warping operation and outputs the warped image for further consumption by deeper networks. The warping operation deals with off-grid points by multi-linear interpolation, hence is differentiable and can back-propagate gradients.

Inspired by FlowNet and STN, we propose {\emphModelNameLongShort}, which enables the unsupervised training of end-to-end CNNs that perform voxel-level 3D medical image registration (see \figureref{flowchart}). The moving image is registered and warped, and the warped image is compared against the fixed image to form a similarity loss. There is a rich body of research into similarity losses \cite{sim-metrics,mi-reg-survey,overlap-inv-mi}. The model is trained to minimize a combination of regularization loss and similarity loss. As the method is unsupervised, the performance potentially increases as it is trained with more unlabeled data. The network consists of several cascaded subnetworks, the number of which might vary, and each subnetwork is responsible for producing a transform that aligns the fixed image and the moving one. Deeper layers register moving images warped according to the output of previous layers with the initial fixed image. The final prediction is the composition of all intermediate flows. It turns out that network cascading significantly improves the performance in the presence of large displacement between the input images. While the idea of cascading subnetworks is found in FlowNet 2.0, our approach does not include as much artificial intervention of network structures as FlowNet 2.0 does. Instead, we employ a natural dichotomy in a subnetwork structure consisting of \emph{affine} and \emph{deformable} registration processes, which is also found in traditional methods, including ANTs (affine and SyN for deformable) \cite{avants2009advanced} and Elastix (affine and B-spline for deformable) \cite{elastix}. Besides these structural innovations, we also introduce the invertibility loss to 3D medical image registration, which encourages backward consistency while achieving better accuracy. Compared with traditional optimization-based algorithms, ours is {\SpeedUpGpu} faster (or {\SpeedUpCpu} faster without GPU acceleration) and achieves state-of-the-art performance.

To summarize, we present a new unsupervised end-to-end learning system using convolutional neural networks for deformation field prediction between 3D medical images. In this framework, we develop 3 technical components:
(1)~We cascade the registration subnetworks, which improves performance for registering largely displaced images without much slow-down;
(2)~We integrate affine registration into our network, which proves to be effective and faster than using a separate tool;
(3)~We incorporate an additional invertibility loss into the training process, which improves registration performance.
The contributions of this work are closely related. An unsupervised approach is very suitable for this problem, as the images are abundant and the ground truth is costly to acquire. The use of the warping operation is crucial to our work, providing the backbone of unsupervision, network cascading and invertibility loss. Network cascading further allows us to plug in different subnetworks, and in this case, the affine registration subnetwork and the deformable ones, enabling us to adapt the natural structure of multiple stages in image registration. The efficient implementation of our algorithm gives a satisfactory speed. The proposed VTN is also used as a building block in \cite{zhao2019recursive}, which thoroughly exploits the idea of deep cascades and the code is publicly available\footnote{\url{https://github.com/microsoft/Recursive-Cascaded-Networks}}.

\section{Related Work}

\subsection{Traditional Algorithms}

Previously, there has been much effort on automating image registration. Tools like FAIR \cite{2009-FAIR}, ANTs \cite{avants2009advanced} and Elastix \cite{elastix} have been developed for automated image registration \cite{oliveira2014medical}. Generally, these algorithms define a space of transformations and a metric of alignment quality, and then find the optimal transformation by iteratively updating the parameters. The optimization process is highly time-consuming, rendering such methods impractical for clinical applications.

The transformation space can be either parametric or non-parametric. Affine transforms can be described by only a few real numbers, whereas a free-form dense deformable field specifies the displacement for each grid point. Though the latter contains all possible transforms, it is common to apply a multi-stage approach to the problem. For example, ANTs \cite{avants2009advanced} registers the input images with a rigid transform, then a general affine transform, and finally a deformable transform modeled by SyN. In this paper, we also have components for affine/deformable registration, which are modeled by neural networks.

Selecting an informative metric is crucial to registration quality. The metrics, also called loss functions, often consist of two parts, one measuring the level of correspondence between the input images implied by the transform, the other regularizing the transform itself. Examples of the former part include photo-metric difference, correlation coefficient and mutual information \cite{mi-reg-survey,overlap-inv-mi} among others \cite{sim-metrics}. Some of these measures, notably mutual information, require binning or quantizing, which makes gradients vanish thus continuous optimization techniques such as gradient descent inapplicable.

\subsection{Supervised Learning Methods}

Lee et al. \cite{cvpr-learn-sim} employ Support Vector Machines (SVM) to learn the similarity measure for multi-modal image registration for brain scans. The training process of their algorithm requires pre-aligned image pairs thus the method is supervised. There have been some works tackling medical image registration with CNNs \cite{cnn-mi-analysis-survey}. Sokooti et al. \cite{sokooti-chest} develop a patch-based CNN to register chest CT scans and trains it with synthetic data. Miao et al. \cite{cnn-rt-2d3d-reg} use CNN to perform 2D/3D registration, in which CNN regressors are used to estimate transformation parameters. Their CNNs are trained over synthetic data and the method is not end-to-end. FlowNet \cite{dosovitskiy2015flownet}, developed by Dosovitskiy et al., is an FCN \cite{long2015fully} for optical flow prediction. The network is trained with synthetic data and estimates pixel-level correspondence.

While supervised learning methods achieve good performance, either abundant groud-truth alignment must be available, or synthetic data are used. Generation of synthetic data has to be carefully designed so that the generated data resemble the real ones.

\subsection{Unsupervised Learning Methods}



In an earlier work towards fast and accurate medical image registration by Shan et al. \cite{flownet-medical-2d-3d}, an unsupervised end-to-end learning-based method for deformable medical image registration is proposed. The method in \cite{flownet-medical-2d-3d} registers 2D images. It is evaluated by registering corresponding sections of MR brain images and CT liver scans. Some later work directly performs 3D registration \cite{de2017end,de2019deep}, fully exploiting the information available from the 3D images. They aim to learn sparse parameters introduced by the traditional algorithms, which probably limits their performance. de Vos et al. \cite{de2019deep} also try to stack multiple networks and investigate an affine network, however, each of them is trained independently and separately. Their method is not end-to-end and only achieves a comparable performance to their baseline methods. \hl{Another work }\cite{shen2019networks}\hl{ proposes to train a multi-step affine network followed by a momentum generator network. Although their framework is claimed to be ``end-to-end'', each stage is still trained separately after fixing the previous ones.} In contrast, with the help of jointly trained cascaded subnetworks, the affine stage is naturally integrated into our framework, rather than done or trained out-of-band in \cite{flownet-medical-2d-3d,de2017end,de2019deep,shen2019networks}.

VoxelMorph, proposed by Balakrishnan et al. \cite{voxelmorph,balakrishnan2019voxelmorph}, is an unsupervised learning-based method for 3D medical image registration that predicts a dense deformation field. Another recently proposed unsupervised method by Krebs et al. \cite{krebs2019learning,krebs2018unsupervised}, based on a low-dimensional probablistic model, performs comparably to VoxelMorph. VoxelMorph contains an encoder-decoder structure, uses warping operation to produce warped moving images, and is trained to minimize the dissimilarity between the warped image and the fixed image. Their method does not consider affine registration and assumes the input images are already affinely aligned, whereas ours embeds the process of affine registration as an integrated part of the network. Furthermore, their algorithm does not work well when large displacement between the images is present, which is common for liver CT scans. Finally, VoxelMorph is designed to take any two images as input, but \cite{voxelmorph} only evaluates it in the atlas-based registration scenario. Consider a clinical scenario where the brain of a patient is captured before and after an operation. It would be better, in terms of accuracy and convenience, to register these two images, instead of registering both to an atlas.

It is noticeable that all the unsupervised methods use the warp operation to train networks without supervision. This paper exploits the operation in a trio (namely enabling unsupervised training, enabling cascading, and implementing invertibility loss, which will be detailed later). We present a more sophisticated and robust design that works well in the presence or absence of large displacement, and we evaluate the methods for general registration among images.

\section{Method}

\subsection{Problem Formulation}

The input of an image registration problem consists of two images $I_{1,2}$, both of which are functions $\Omega\to\mathbb{R}^c$, where $\Omega$ is a region of $\mathbb{R}^n$ and $c$ denotes the number of channels. Since this work focuses on 3D medical image registration, we confine ourselves to the case where $\Omega$ is a cuboid and $c=1$ (grayscale image). Specifically, this means $\Omega\subseteq\mathbb{R}^3$ and each image is a function $\Omega\to\mathbb{R}$. The objective of image registration is to find a displacement field (or flow field) $f_{12}:\Omega\to\mathbb{R}^3$ so that
\begin{equation}
I_1\left(x\right)\approx I_2\left({x+f\left(x\right)}\right),
\end{equation}
where the precise meaning of ``$\approx$'' depends on specific application. The field $f_{12}$ is called the flow from $I_1$ to $I_2$ since it tells where each voxel in $I_1$ is in $I_2$. We define $\Warp{I_2}{f}$ as the image $I_2$ warped according to $f$, i.e., $\Warp{I_2}{f}\left(x\right)=I_2\left({x+f\left(x\right)}\right)$. The above objective can be rephrased as finding $f$ maximizing the similarity between $I_1$ and $\Warp{I_2}{f}$.

The image $I_1$ is also called the fixed image, and $I_2$ the moving one. The term ``moving'' suggests that the image is transformed during the registration process.

Consider warping an image twice, first with $g_1$ then with $g_2$. What this procedure produces is
\begin{align}
\label{comp-flow-eq}
&\Warp{\Warp{I}{g_1}}{g_2}\left(x\right)
\nonumber\\
{}={}&\Warp{I}{g_1}\left({x+g_2\left(x\right)}\right)
\nonumber\\
{}={}&I\left({x+g_2\left(x\right)+g_1\left({x+g_2\left(x\right)}\right)}\right)
\nonumber\\
{}={}&\Warp{I}{g_2+\Warp{g_1}{g_2}}\left(x\right).
\end{align}
This motivates the definition of the \emph{composition} of two flows. If we define the composition of the flow fields $g_1,g_2$ to be
\begin{equation}
\label{comp-flow-def}
g_1\star g_2=g_2+\Warp{g_1}{g_2},
\end{equation}
\equationref{comp-flow-eq} can be restated as
\begin{equation}
\Warp{\Warp{I}{g_1}}{g_2}
=
\Warp{I}{g_1\star g_2}.
\end{equation}

It is noticeable that the warp operation in the above formulation should be further specified in practice. Real images as well as flow fields are only defined on lattice points. We continuate them onto the enclosing cuboid by trilinear interpolation as done in \cite{jaderberg2015spatial}. Furthermore, we deal with out-of-bound indices by nearest-point interpolation. That is, to evaluate a function defined on lattice points at any point $x$, we first move $x$ to the nearest in the enclosing cuboid of those lattice points, then interpolate the value from the 8 nearest lattice points.

\subsection{Unsupervised End-to-End Registration Network}

Our network, called {\emphModelNameLongShort}, consists of several cascaded registration subnetworks, after each of which the moving image is warped. The unsupervised training of network parameters is guided by the dissimilarity between the fixed image and each of the warped images, with the regularization losses on the flows predicted by the subnetworks.

\hl{The warping operation, also known as the sampler in STN }\cite{jaderberg2015spatial}\hl{, is differentiable to both the input image and the input flow field due to the trilinear interpolation. Those warping operations can back-propagate gradients to all the preceding subnetworks, which is critical for the end-to-end learning of our cascaded networks.}

In deformable image registration, it is common to apply an initial rigid transformation as a global alignment before predicting the dense flow field. Instead of prepending a time-consuming preprocessing stage with a tool like ANTs \cite{avants2009advanced} as done in VoxelMorph \cite{voxelmorph}, we integrate this procedure as a top-level subnetwork. \hl{Our affine registration subnetwork predicts a set of affine parameters, after which a flow field can be generated for warping.} The integrated affine registration subnetwork not only works in negligible running time, but also outperforms the traditional affine stage.

\subsection{Loss Functions}

To train our model in an unsupervised manner, we measure the (dis)similarity between the moving images warped by the spatial transformer and the fixed image. 
Regularization losses are introduced to prevent the flow fields from being unrealistic or overfitting. \hl{We use correlation coefficient as the similarity measurement, and the total variation loss as the regularization term for dense flow predictions. Furthermore, we introduce orthogonality loss and determinant loss as the regularization terms for the affine registration subnetwork, which are essential in preventing the gradients from exploding and ensuring training stability. Those loss functions are discussed as follows.}

\paragraph{Correlation Coefficient} The covariance between $I_1$ and $I_2$ is defined as
\begin{equation}
\COVAR{I_1}{I_2}=
\frac{1}{\left|\Omega\right|}
\sum_{x\in\Omega}{I_1\left(x\right)I_2\left(x\right)}
-
\frac{1}{{\left|\Omega\right|}^2}
\sum_{x\in\Omega}{I_1\left(x\right)}
\sum_{y\in\Omega}{I_2\left(y\right)},
\end{equation}
where $\Omega$ denotes the cuboid (or grid) on which the input images are defined. Their correlation coefficient is defined as
\begin{equation}
\Corr{I_1}{I_2}=\frac{\COVAR{I_1}{I_2}}{\sqrt{\COVAR{I_1}{I_1}\COVAR{I_2}{I_2}}}.
\end{equation}
The images are regarded as random variables whose sample space is the points on which voxel values are available. The range of correlation coefficient is $\left[{-1,1}\right]$, it measures how much the two images are linear related, and attains $\pm1$ if and only if the two are linear function of each other. Applying a non-degenerate linear function to any of the images does not change their correlation coefficient, therefore, this measure is more robust than $L_2$ loss. For real-world images, the correlation coefficient should be non-negative (unless one of the images is a negative film). The correlation coefficient loss is defined as
\begin{equation}
L_\CorrCoef\left({I_1,I_2}\right)=1-\Corr{I_1}{I_2}.
\end{equation}

\paragraph{Total Variation Loss (Smooth Term)} For a dense flow field, we regularize it with the following loss that discourages discontinuity:
\begin{equation}
L_\TV = \frac{1}{3|\Omega|}\sum_{x}\sum_{i=1}^{3} { (f(x+e_i) - f(x))^2},
\end{equation}
where $e_{1,2,3}$ form the natural basis of $\mathbb{R}^3$. \hl{This varies from the initial definition of the total variation loss }\cite{rudin1992nonlinear}\hl{ (which includes a square root), but our formula is more natural as a loss term (referring to the L2 regularization).}

\paragraph{Orthogonality Loss} For the specific task discussed in this paper (medical image registration), it is usually the case that the input images need only a small scaling and a rotation before they are affinely aligned. We would like to penalize the network for producing overly non-rigid transform. To this end, we introduce a loss on the non-orthogonality of $I+A$, where \hl{$I$ denotes the identity matrix} and $A$ denotes the transform matrix produced by the affine registration network (see Section~\ref{ssec_affine} for more details). Let $\lambda_{1,2,3}$ be the singular values of $I+A$, the orthogonality loss is
\begin{equation}
L_\Orthogonal=-6+\sum_{i=1}^{3}{\left({\lambda_i^2+\lambda_i^{-2}}\right)}.
\end{equation}
\hl{The motivation of this formula is mainly that a matrix is orthogonal if and only if all its singular values are $1$. Hence, the more deviant $I+A$ is from being an orthogonal matrix (i.e., the more its singular values deviating from ones), the larger its orthogonality loss. If $I+A$ is orthogonal, the value will be zero.}

\hl{Computing orthogonality loss involves singular values of $I+A$. The square of those singular values are exactly the eigenvalues of ${\left({I+A}\right)}^\Transpose{\left({I+A}\right)}$. Since the loss is a symmetric function of those eigenvalues, it can be rewritten as a fraction w.r.t.\@ the coefficients of the characteristic polynomial of ${\left({I+A}\right)}^\Transpose{\left({I+A}\right)}$ by Vi{\`e}te's theorem. Then the derivatives could be directly calculated.}

\paragraph{Determinant Loss} We assume images are taken with the same chirality, therefore, an affine transform involving reflection is not allowed. This imposes the requirement that $\det\left({I+A}\right)>0$. Together with the orthogonality requirement, we set the determinant loss to be
\begin{equation}
L_{\det}={\left({-1+\det\left({A+I}\right)}\right)}^2,
\end{equation}
\hl{where $I$ denotes the identity matrix and $A$ denotes the transform matrix produced by the affine registration network.}

\section{Network Architecture}

\subsection{Cascading}

Each subnetwork is responsible for aligning the fixed image and the \emph{current} moving image. Following each subnetwork, the moving image is warped with the predicted flow, and the warped image is fed into the next cascaded subnetwork. The flow fields are composed to produce the final estimation. \figureref{wholenet} illustrates how the networks are cascaded, how the images are transformed and how each part contributes to the loss. \hl{With all layers being differentiable, gradients can back-propagate to the whole system and enable the unsupervised end-to-end learning.}

It might be tempting to compare our scheme with that of FlowNet 2.0 \cite{IMKDB17}. FlowNet 2.0 stacks subnetworks in a different way than our method. It performs two separate lines of flow estimations (large/small displacement flows) and fuses them into the final estimation. Each of its intermediate subnetworks has inputs of not only the warped moving image and the fixed image, but also the initial moving image, the current flow, and the brightness error. Its subnetworks are carefully crafted, having similar yet different structures and expected to solve specific problems (e.g., large/small displacement) in flow estimation. In contrast, our method does not involve two separate lines of registration, i.e., each subnetwork works on the fixed image and the warped moving image produced by the previous one, and intermediate subnetworks do not get more input than the initial one. We do not interfere much with the structures of subnetworks. Despite the initial affine subnetwork, we do not assign specific tasks to the remaining subnetworks since they share the same structure.

\subsection{Affine and Dense Deformable Subnetworks} \label{ssec_affine}

\AffineNetFigure{tbp}
\DenseNetFigure{tbp}

The affine registration subnetwork aims to align the input image with an affine transform. It is only used as our first subnetwork. As illustrated in \figureref{affinenet}, the input is downsampled by strided 3D convolutions, and finally a fully-connected layer is applied to produce $12$ numeric parameters as output, which represents a $3\times 3$ transform matrix $A$ and a $3$-dimensional displacement vector $b$. As a common practice, the number of channels doubles as the length of resolution halves. The flow field produced by this subnetwork is defined as
\begin{equation}
\label{affine-flow-def}
f\left(x\right)=Ax+b.
\end{equation}
\hl{The moving image is then transformed according to the output affine parameters and fed into the subsequent deformable registration subnetworks.}

The dense deformable registration subnetwork is used as all subsequent subnetworks, each of which refines the registration based on the output of the subnetwork preceeding it. It follows an encoder-decoder architecture, as illustrated in \figureref{densenet}, which is commonly used for dense prediction. We use strided 3D convolution to progressively downsample the image, and then use deconvolution (transposed convolution) \cite{long2015fully} to recover spatial resolution. \hl{As suggested in U-Net }\cite{ronneberger2015u}\hl{, skip connections between the convolutional layers and the deconvolutional layers are added to help refining dense prediction.} The subnetwork will output the dense flow field, a volume feature map with $3$ channels ($x,y,z$ displacements) of the same size as the input.

\subsection{Invertibility}

\InvLossFigure{t}

Given two images $I_{1,2}$, going from a voxel in $I_1$ to the corresponding voxel in $I_2$ then back to $I_1$ should give zero displacement. Otherwise stated, the registration should be \emph{round-trip}. In \figureref{invloss-fig}, we demonstrate the possible situations. The pair of solid arrows exemplifies round-trip registration, whereas the pair of dashed arrows exemplifies non-round-trip registration. If we have computed two flow fields (back and forth), $f_{12}$ and $f_{21}$, the composed fields exhibit the round-trip behavior of the registration, as illustrated by the magenta straight arrow in \figureref{invloss-fig}. Ideally, round-trip registration should satisfy the equations $f_{12}\star f_{21}=f_{21}\star f_{12}=0$. We capture the round-tripness for a pair of images with the invertibility loss, namely
\begin{equation}
L_\Inv = {\left\|{f_{12}\star f_{21}}\right\|}_2^2 + {\left\|{f_{21}\star f_{12}}\right\|}_2^2.
\end{equation}
The larger the invertibility loss, the less round-trip the registration. For perfectly round-trip registration, the invertibility loss is zero. We come up with, formulate, and implement the invertibility loss independently of \cite{godard2017unsupervised}. We use L2 invertibility loss whereas \cite{godard2017unsupervised} uses L1 left-right disparity consistency loss, which is just a matter of choice. We are the first to incorporate the invertibility loss into 3D images to boost performance on medical image tasks.

\section{Experiment}

We evaluate our algorithm with extensive experiments on both liver CT datasets and brain MRI datasets. \hl{The models are trained separately for the two types of images. We train and test the models for pairwise, subject-to-subject registration. This setting is more general than the atlas-based registration as studied in VoxelMorph }\cite{voxelmorph}\hl{, where all moving images are assumed to be registered to the fixed atlas.}

\hl{The input images to the models are assumed to be of size $128^3$. We apply spatial augmentation during training, where the input images are transformed by random B-Spline }\cite{rueckert1999nonrigid}\hl{ fields of $5 \times 5 \times 5$ control points within a maximum displacement of $12$.}

We compare our algorithm against state-of-the-art traditional registration algorithms including ANTs\footnote{\hl{The command we use for ANTs:}
\texttt{-d 3 -o <OutFileSpec>
-u 1 -w [0.025,0.975]
-r [<Fixed>,<Moving>,1]
-t Rigid[0.1]
-m MI[<Fixed>,<Moving>,1,32,Regular,0.2]
-c [2000x2000x2000,1e-9,15]
-s 2x1x0 -f 4x2x1
-t Affine[0.1]
-m MI[<Fixed>,<Moving>,1,32,Regular,0.1]
-c [2000x2000x2000,1e-9,15]
-s 2x1x0 -f 4x2x1
-t SyN[0.15,3.0,0.0]
-m CC[<Fixed>,<Moving>,1,4]
-c [100x100x100x50,1e-9,15]
-s 3x2x1x0 -f 6x4x2x1}} \cite{avants2009advanced} and Elastix\footnote{\hl{The command we use for Elastix:}
\texttt{-f <Fixed> -m <Moving>
-out <OutFileSpec>
-p Affine -p BSpline{\textunderscore}1000}} \cite{elastix}, as well as VoxelMorph \cite{voxelmorph}. Our algorithm achieves state-of-the-art performance while being much faster. Our experiments prove that the performance of our unsupervised method is improved as more unlabeled data are used in training. We show that cascading subnetworks significantly improves the performance, and that integrating affine registration into the method is effective.

We evaluate the performance of algorithms with the following metrics:
\begin{itemize}
\item\textbf{Seg. IoU} is the Jaccard coefficient between the warped liver segmentation and the ground truth. We warp the segmentation of the moving image by the predicted deformable field, and compute the Jaccard coefficient of the warped segmentation with the ground-truth segmentation of the fixed image. ``IoU'' means ``intersection over union'', i.e., $\frac{\left|{A\cap B}\right|}{\left|{A\cup B}\right|}$, where $A,B$ are the set of voxels the organ consists of.
\item\textbf{Lm. Dist.} is the average distance between warped landmarks (points of anatomical interest) and the ground truth.
\item\textbf{Time} is the average time taken for each pair of images to be registered. Some methods are implemented with GPU acceleration, therefore there are two versions of this metric (with or without GPU acceleration).
\end{itemize}

Our model is defined and trained using TensorFlow \cite{tensorflow}. We accelerate training with nVIDIA TITAN Xp and CUDA 8.0. We use the Adam optimizer \cite{adam-optimizer} with the default parameters in TensorFlow 1.4. The batch size is 8 pairs per batch. The initial learning rate is ${10}^{-4}$ and halves every epoch after the $4^{\mathrm{th}}$ epoch. Each epoch consists of 20000 batches and the number of epochs is 5. Performance evaluation uses the same GPU model. Traditional methods work on CPU. We also test neural-network-based methods with GPU acceleration disabled for a fairer comparison of speed. The CPU model used is Intel\textsuperscript{\textregistered} Xeon\textsuperscript{\textregistered} CPU E5-2690 v4 @ 2.60GHz (14 Cores).

\subsection{Experiments on Liver Datasets}

\subsubsection{Settings}\label{expr-liver-params}

The input to the algorithms are liver CT scans of size ${128}^3$. Affine subnetworks downsample the images to $4^3$ before applying the fully-connected layer. Dense deformable subnetworks downsample the images to $2^3$ before doing transposed deconvolution.

We cascade \emph{up to} 4 registration subnetworks. The reason we need to cascade multiple subnetworks is that large displacement is very common among CT liver scans and that with more subnetworks, images with large displacement can be progressively aligned. Among these networks, the one with one affine registration subnetwork and three dense deformable registration networks (referred to as ``ADDD'') is used to be trained with different amount of data and compared with other algorithms. We use the correlation coefficient as our similarity loss, the orthogonality loss and the determinant loss as regularization losses for the affine subnetwork, and the total variation loss as that for dense deformable subnetworks. The ratio of losses for ``ADDD'' is listed in \tableref{loss-ratio-table}. The performance is not very sensitive to the choice of hyper-parameters, since each of the dense deformable subnetworks can automatically learn to progressively align the images, and only the final subnetwork and the affine subnetwork need to be trained with similarity loss.

\LossRatioTable{bhtp}

\subsubsection{Datasets}

We have three datasets available:
\begin{itemize}
\item\textbf{LITS} \cite{lits-data} consists of 130 volumes. LITS comes with segmentation of liver, but we do not use such information. This dataset is used for training.
\item\textbf{BFH} is provided by Beijing Friendship Hospital and consists of 92 volumes. This dataset is used for training.
\item\textbf{MICCAI} (MICCAI'07) \cite{miccai-data} consists of 20 volumes with liver segmentation ground truth. We choose 4 points of anatomical interest as the landmarks\footnote{The landmarks: (L1)~the top point of hepatic portal; (L2)~the intersection of the superior and anteroir branches of the right lobe; (L3)~the intersection of the superior and inferior branches of the right lobe; and (L4)~the intersection of the medial and inferior branches of the left lobe.} and ask 3 expert doctors to annotate them, taking their average as the ground truth. This dataset is used as the test data.
\end{itemize}
\hl{All pairs of images in the mixture of LITS and BFH are sampled into mini-batches for training. All image pairs in MICCAI are tested during evaluation.} We crop raw liver CT scans to a volume of size ${128}^3$ around the liver, and normalize them by adjusting exposure so that the histograms of the images match each other. The preprocessing stage is necessary as the images come from different sources.

\LiverMethodsFigure{t}

\LiverCascadeFigure{t}

\subsubsection{Comparison among Methods}\label{comp-trads-sec}

In \tableref{liver-algo-table}, ``ADDD'' is our model detailed in \sectionref{expr-liver-params}, and ``ADDD + inv'' is that model trained with additional term of invertibility loss in the central area (the beginning and the ending quaters of each side are removed) with relative weight ${10}^{-3}$. Learning-based methods ({\ModelNameShort} and VoxelMorph) are trained on LITS and BFH datasets. All methods are evaluated on the MICCAI dataset. To prove the effectiveness of our unsupervised method, we also train ``ADDD'' supervised (the row ``supervised''), where the output of ANTs is used as the ground truth (using end-point error \cite{dosovitskiy2015flownet} plus regularization term as the loss function).

\LiverMethodsTable{hptb}

VoxelMorph-2 is trained (using code released by \cite{voxelmorph}) with a batch size of 4 and an iteration count of 5000. Its performance with a batch size of 8 or 16 is worse so a batch size of 4 is used. After 5000 iterations, the model begins to overfit. The reason that it does not perform well on liver datasets might be that it is designed for brain registration thus cannot handle large displacement.

The results in \tableref{liver-algo-table} show the vast speed-up of learning-based methods against optimization-based methods. \hl{The Wilcoxon signed-rank test indicates that our methods significantly surpass state-of-the-art registration algorithms in terms of both Segmentation IoU and Landmark Distance.}

\tableref{liver-jac-table}\hl{ presents the smoothness metrics (as done in }\cite{de2019deep}\hl{) of our method compared to the traditional methods. It is not surprising that our significantly better performance comes at the price of worse deformation smoothness, but the fraction of folding is still well acceptable (less than 1\%). The invertibility loss has a considerable impact on reducing the fraction of folding.}

\LiverJacTable{hptb}

\figureref{liver-algo-fig} compares different methods listed in \tableref{liver-algo-table}, where three landmarks are selected and the sections of the volumes at the height of each landmark \emph{in the fixed image} are rendered. This means the red crosses (landmarks in the moving and warped images) indicate the projections of the landmarks onto those planes. It should be noted that though the sections of the warped segmentations seem to be less overlapping with those of the fixed one, the Segmentation IoU is computed for the volume and not the sections. It might well be the case that the overlap is not so satisfactory when viewed from those planes yet is better when viewed as a volume. Similarly, overlapping red and yellow crosses do not necessarily imply overlapping fixed and warped landmarks as they might deviate along $z$-axis.

\hl{In }\figureref{liver-cascade-fig}\hl{, we compare ADDD + inv, ADDD, ADD, AD and D. It shows that network cascading better aligns the images with the presence of large displacement, while the invertibility loss has a remarkable effect on the liver boundary.}

\subsubsection{Performance with Different Amount of Data}

In \tableref{liver-moredata-table}, the ``ADDD'' network (see \sectionref{expr-liver-params}) \hl{trained with full data (LITS + BFS) is compared with that trained with only a part of data.} The result demonstrates that training with more \emph{unlabeled} data improves the performance. Since we do not need any annotations during the training phase, there are abundant clinical data ready for use.

\LiverMoreDataTable{hptb}

\subsubsection{Network Cascading}

In \tableref{liver-cascade-table}, all networks are trained on LITS + BFH. The models whose name does not include ``A'' have the affine registration subnetwork removed, and the number of ``D''s is the number of dense deformable registration subnetworks. (See \sectionref{expr-liver-params} for ``ADDD''.)

\LiverCascadeTable{htbp}

\subsection{Experiments on Brain Datasets}

\paragraph{Settings} The input to the algorithms are brain MR images of size ${128}^3$. For some experiments, the selection of which will be detailed later, the brain scans are preprocessed to be aligned to a specific atlas from LONI \cite{loni}. We use ANTs for this purpose.
There are two reasons we align the brain scans with ANTs. One is that VoxelMorph \cite{voxelmorph} requires the input to be affinely registered. The other is that we will compare the performance between ``ANTs + deformable registration subnetworks'' and ``affine registration subnetwork + deformable registration subnetworks'', i.e., to compare the effectiveness of integrating our affine registration subnetwork in place of ANTs.

The following methods will have ANTs-affine-aligned images as input: VoxelMorph methods and {\ModelNameShort} without ``A'' (i.e., ``D'', ``DD'' and ``DDD''). A more precise naming is ``ANTs + VoxelMorph'' and ``ANTs + {\ModelNameShort}''. The following methods will \emph{not} have ANTs-affine-aligned images as input: ANTs, Elastix, {\ModelNameShort} with ``A'' (i.e., ``AD'', ``ADD'' and ``ADDD''). Those methods have affine registration integrated. The comparison inside the former group focuses on dense deformable registration performance, that inside the latter group on overall performance, and that among the two groups benchmarks affine registration subnetwork versus ANTs affine registration in the context of a complete registration pipeline.

We will show 3 sets of comparisons similar to those for liver datasets. In the tables listed in later paragraphs, the time (with or without GPU) does not include the preprocessing with ANTs even if it is used. Preprocessing an image with ANTs costs 73.94 seconds on average. We will mention this fact again when such emphasis is needed.

Care should be taken when evaluating methods with ANTs affine alignment. For the data to be comparable with those with affine registration integrated, the fixed image should be equivalent. Methods with ANTs affine alignment have both moving and fixed images aligned to an atlas. Those with integrated affine registration never move the fixed image. The affine transform produced by ANTs might not be orthogonal, which is the source of unfair comparison. If the affine transform is shrinking, methods with ANTs affine alignment gain advantage. Otherwise, methods with integrated affine alignment do.

One measure, Segmentation IoU, is not affected, because the volumes of all objects get multiplied by the determinant of the affine transform and the evaluation measure is homogeneous. For Landmark Distance, we perform the inverse of the linear part of the affine transform (which aligns the fixed image to the atlas) to the difference vector between warped landmark and landmark in the (aligned) fixed image, so that the length goes back to the coordinate defined by the original fixed image. This way, we minimize loss of precision to prevent unfairly underevaluating methods with ANTs affine alignment. Speaking of the actual data, the affine transformations produced by ANTs are slightly shrinking. Our correction restores a fair comparison among all methods.

\paragraph{Datasets} We use volumes from the following datasets for training:
\begin{itemize}
\item ADNI \cite{adni} (67 volumes);
\item ABIDE-1 \cite{abide} (318 volumes): part of data from ABIDE;
\item ABIDE-2 \cite{abide} (1101 volumes): the rest from ABIDE;
\item ADHD \cite{adhd} (973 volumes).
\end{itemize}
We acquire the second part of ABIDE after a while when the first part was downloaded and processed, thus the split. This only helps us to understand how performance improves as more data are used for training. For comparison among different methods, it is always the case that all the data mentioned above are used for training. \hl{All pairs of images in the mixture of those datasets are sampled into mini-batches for training.}

Raw MR scans are cropped to ${128}^3$ around the brain. The scans are normalized based on the histograms of their foreground color distribution, which might vary because they are captured on different sites.

For evaluation, we use 20 volumes from the LONI Probabilistic Brain Atlas (LPBA40) \cite{loni}. LONI consists of 40 volumes, 20 of which have tilted head positions and are discarded. For the remaining 20 volumes, 18 landmarks\footnote{The landmarks: (L1)~right lateral ventricle posterior; (L2)~left lateral ventricle posterior; (L3)~anterior commissure corresponding to the midpoint of decussation of the anterior commissure on the coronal AC plane; (L4)~right lateral ventricle superior; (L5)~right lateral ventricle inferior; (L6)~left lateral ventricle superior; (L7)~left lateral ventricle inferior; (L8)~middle of lateral ventricle; (L9)~posterior commissure corresponding to the midpoint of decussation; (L10)~right lateral ventricle superior; (L11)~left lateral ventricle superior; (L12)~middle of lateral ventricle; (L13)~corpus callosum inferior; (L14)~corpus callosum superior; (L15)~corpus callosum anterior; (L16)~corpus callosum posterior tip of genu corresponding to the location of the most posterior point of corpus callosum posterior tip of genu on the midsagittal planes; (L17)~corpus callosum fornix junction; and (L18)~pineal body.} are annotated by 3 experts and the average are taken as the ground truth.

\BrainMethodsFigure{t}

\paragraph{Comparison Among Methods} In \tableref{brain-algo-table}, we compare different methods on brain datasets. All neural networks are trained on all available training data, i.e., ADNI, ABIDE and ADHD. In the table, ``supervised'' is our ``ADD'' model supervised with ANTs as the ground truth. Its loss is the end-point error \cite{dosovitskiy2015flownet}. \tableref{brain-jac-table}\hl{ presents the smoothness metrics of our method compared to the traditional methods.}

\BrainMethodsTable{hptb}

\BrainJacTable{hptb}

Among these methods, our ``ADD'' achieves the lowest Landmark Distance with a competitive speed. \hl{If we compare ``ADD'' with ``DD'', the Wilcoxon signed-rank test indicates that the integration of affine registration subnetwork significantly improves the Landmark Distance, compared to using ANTs for out-of-band affine alignment. Although there is still a performance gap compared to ANTs in terms of Segmentation IoU, it is remarkable that our method generates significantly less folding area.}

\figureref{brain-algo-fig} exemplifies the methods on three pairs of scans. Comparison between our methods and traditional methods proves the applicability of our methods to 3D brain registration. Comparison between ADD and {DD\xtraAntsAffine} shows that integrating affine registration subnetwork is effective.

\paragraph{Performance with Different Amount of Data} In \tableref{brain-moredata-table}, we summarize the performance of ``DD'' (with ANTs affine alignment) trained on different amount of unlabeled data. As more data are used to train the network, its performance in terms of Landmark Distance consistently increases.

\BrainMoreDataTable{hptb}

\paragraph{Network Cascading and Integration of Affine Registration} \tableref{brain-cascade-table} compares the performances of differently cascaded networks. The networks without ``A'' have ANTs-affine-aligned images as input, whereas the networks with ``A'' do not.

\BrainCascadeTable{hptb}

As one would expect, the performance in each group improves as the model gets more levels of cascaded subnetworks. While the methods with ANTs affine alignment have higher Segmentation IoU, integrating affine registration subnetwork yields better Landmark Distance. Worth mentioning is that the better Segmentation IoU comes at the price of a rather slow preprocessing phase (74 seconds).

\section{Discussion}

\hl{Our method achieves significant performance gain over traditional methods on the liver CT dataset, however, some method (ANTs) still performs slightly better on the brain MRI dataset. This is probably because brain MRIs are mainly of small displacements which might be better suited for traditional iterative methods. Besides the correlation coefficient loss used in this paper, other similarity measurements like cross correlation can also be explored for better accuracy.}

\section{Conclusion}

In this paper, we present {\emphModelNameLongShort}, a new unsupervised end-to-end learning framework using convolutional neural networks for 3D medical image registration. The network is trained in an unsupervised manner without any ground-truth deformation. Experiments demonstrate that our method achieves state-of-the-art performance, and that it witnesses an {\SpeedUpGpu} (or {\SpeedUpCpu} without GPU acceleration) speed-up compared to traditional medical image registration methods. Our thorough experiments prove our contributions, each on its own being useful and forming a strong union when put together. Our networks can be cascaded. Cascading deformable subnetworks tackles the difficulty of registering images in the presence of large displacement. Network cascading also enables the integration of affine registration into the algorithm, resulting in a truly end-to-end method. The integration proves to be more effective than using out-of-band affine alignment. We also incorporate the invertibility loss into the training process, which further enhances the performance. Our methods can potentially be applied to various other medical image registration tasks.

\section*{Acknowledgment}

The authors would like to thank Beijing Friendship Hospital for providing the ``BFH'' dataset as well as other data providers for making their data publicly available.

\ifCLASSOPTIONcaptionsoff
  \newpage
\fi

\bibliographystyle{IEEEtran}
\bibliography{IEEEabrv,refs}

\begin{thebibliography}{10}
\providecommand{\url}[1]{#1}
\csname url@samestyle\endcsname
\providecommand{\newblock}{\relax}
\providecommand{\bibinfo}[2]{#2}
\providecommand{\BIBentrySTDinterwordspacing}{\spaceskip=0pt\relax}
\providecommand{\BIBentryALTinterwordstretchfactor}{4}
\providecommand{\BIBentryALTinterwordspacing}{\spaceskip=\fontdimen2\font plus
\BIBentryALTinterwordstretchfactor\fontdimen3\font minus
  \fontdimen4\font\relax}
\providecommand{\BIBforeignlanguage}[2]{{%
\expandafter\ifx\csname l@#1\endcsname\relax
\typeout{** WARNING: IEEEtran.bst: No hyphenation pattern has been}%
\typeout{** loaded for the language `#1'. Using the pattern for}%
\typeout{** the default language instead.}%
\else
\language=\csname l@#1\endcsname
\fi
#2}}
\providecommand{\BIBdecl}{\relax}
\BIBdecl

\bibitem{oliveira2014medical}
F.~P.~M. Oliveira and J.~M. R.~S. Tavares, ``Medical image registration: a
  review,'' \emph{Computer Methods in Biomechanics and Biomedical Engineering},
  vol.~17, no.~2, pp. 73--93, 2014.

\bibitem{Hermosillo2002}
\BIBentryALTinterwordspacing
G.~Hermosillo, C.~Chefd'Hotel, and O.~Faugeras, ``Variational methods for
  multimodal image matching,'' \emph{International Journal of Computer Vision},
  vol.~50, no.~3, pp. 329--343, Dec 2002. [Online]. Available:
  \url{https://doi.org/10.1023/A:1020830525823}
\BIBentrySTDinterwordspacing

\bibitem{mi-reg-freeform}
\BIBentryALTinterwordspacing
X.~Huang, N.~Paragios, and D.~N. Metaxas, ``Shape registration in implicit
  spaces using information theory and free form deformations,'' \emph{IEEE
  Trans. Pattern Anal. Mach. Intell.}, vol.~28, no.~8, pp. 1303--1318, Aug.
  2006. [Online]. Available: \url{https://doi.org/10.1109/TPAMI.2006.171}
\BIBentrySTDinterwordspacing

\bibitem{avants2009advanced}
B.~B. Avants, N.~Tustison, and G.~Song, ``Advanced normalization tools
  (ants),'' \emph{Insight j}, vol.~2, pp. 1--35, 2009.

\bibitem{elastix}
S.~Klein, M.~Staring, K.~Murphy, M.~A. Viergever, and J.~P.~W. Pluim,
  ``elastix: A toolbox for intensity-based medical image registration,''
  \emph{IEEE Transactions on Medical Imaging}, vol.~29, no.~1, pp. 196--205,
  Jan 2010.

\bibitem{dosovitskiy2015flownet}
A.~Dosovitskiy, P.~Fischery, E.~Ilg, P.~Hausser, C.~Hazirbas, V.~Golkov, P.~V.
  Der~Smagt, D.~Cremers, and T.~Brox, ``Flownet: Learning optical flow with
  convolutional networks,'' \emph{International Conference on Computer Vision},
  pp. 2758--2766, 2015.

\bibitem{IMKDB17}
\BIBentryALTinterwordspacing
E.~Ilg, N.~Mayer, T.~Saikia, M.~Keuper, A.~Dosovitskiy, and T.~Brox, ``Flownet
  2.0: Evolution of optical flow estimation with deep networks,'' in \emph{IEEE
  Conference on Computer Vision and Pattern Recognition (CVPR)}, 2017.
  [Online]. Available:
  \url{http://lmb.informatik.uni-freiburg.de/Publications/2017/IMKDB17}
\BIBentrySTDinterwordspacing

\bibitem{long2015fully}
J.~{Long}, E.~{Shelhamer}, and T.~{Darrell}, ``Fully convolutional networks for
  semantic segmentation,'' in \emph{2015 IEEE Conference on Computer Vision and
  Pattern Recognition (CVPR)}, 2015, pp. 3431--3440.

\bibitem{stn}
\BIBentryALTinterwordspacing
M.~Jaderberg, K.~Simonyan, A.~Zisserman, and k.~kavukcuoglu, ``Spatial
  transformer networks,'' in \emph{Advances in Neural Information Processing
  Systems 28}, C.~Cortes, N.~D. Lawrence, D.~D. Lee, M.~Sugiyama, and
  R.~Garnett, Eds.\hskip 1em plus 0.5em minus 0.4em\relax Curran Associates,
  Inc., 2015, pp. 2017--2025. [Online]. Available:
  \url{http://papers.nips.cc/paper/5854-spatial-transformer-networks.pdf}
\BIBentrySTDinterwordspacing

\bibitem{sim-metrics}
\BIBentryALTinterwordspacing
D.~J.~H. A.~Melbourne, G.~Ridgway, ``Image similarity metrics in image
  registration,'' pp. 7623 -- 7623 -- 10, 2010. [Online]. Available:
  \url{https://doi.org/10.1117/12.840389}
\BIBentrySTDinterwordspacing

\bibitem{mi-reg-survey}
J.~P.~W. Pluim, J.~B.~A. Maintz, and M.~A. Viergever,
  ``Mutual-information-based registration of medical images: a survey,''
  \emph{IEEE TRANSCATIONS ON MEDICAL IMAGING}, pp. 986--1004, 2003.

\bibitem{overlap-inv-mi}
C.~Studholme and D.~L.~G. Hill, ``D.j.hawkes. an overlap invariant entropy
  measure of 3d medical image alignment,'' \emph{Pattern Recognition}, no.~32,
  1999.

\bibitem{zhao2019recursive}
S.~Zhao, Y.~Dong, E.~I.-C. Chang, and Y.~Xu, ``Recursive cascaded networks for
  unsupervised medical image registration,'' in \emph{The IEEE International
  Conference on Computer Vision (ICCV)}, October 2019.

\bibitem{2009-FAIR}
J.~Modersitzki, \emph{{FAIR}: Flexible Algorithms for Image
  Registration}.\hskip 1em plus 0.5em minus 0.4em\relax Philadelphia: SIAM,
  2009.

\bibitem{cvpr-learn-sim}
D.~Lee, M.~Hofmann, F.~Steinke, Y.~Altun, N.~D. Cahill, and B.~Scholkopf,
  ``Learning similarity measure for multi-modal 3d image registration,'' in
  \emph{2009 IEEE Conference on Computer Vision and Pattern Recognition}, June
  2009, pp. 186--193.

\bibitem{cnn-mi-analysis-survey}
\BIBentryALTinterwordspacing
G.~J.~S. Litjens, T.~Kooi, B.~E. Bejnordi, A.~A.~A. Setio, F.~Ciompi,
  M.~Ghafoorian, J.~A. W.~M. van~der Laak, B.~van Ginneken, and C.~I.
  S{\'{a}}nchez, ``A survey on deep learning in medical image analysis,''
  \emph{CoRR}, vol. abs/1702.05747, 2017. [Online]. Available:
  \url{http://arxiv.org/abs/1702.05747}
\BIBentrySTDinterwordspacing

\bibitem{sokooti-chest}
H.~Sokooti, B.~de~Vos, F.~Berendsen, B.~P.~F. Lelieveldt, I.~I{\v{s}}gum, and
  M.~Staring, ``Nonrigid image registration using multi-scale 3d convolutional
  neural networks,'' in \emph{Medical Image Computing and Computer Assisted
  Intervention − MICCAI 2017}, M.~Descoteaux, L.~Maier-Hein, A.~Franz,
  P.~Jannin, D.~L. Collins, and S.~Duchesne, Eds.\hskip 1em plus 0.5em minus
  0.4em\relax Cham: Springer International Publishing, 2017, pp. 232--239.

\bibitem{cnn-rt-2d3d-reg}
S.~Miao, Z.~J. Wang, and R.~Liao, ``A cnn regression approach for real-time
  2d/3d registration,'' \emph{IEEE Transactions on Medical Imaging}, vol.~35,
  no.~5, pp. 1352--1363, May 2016.

\bibitem{flownet-medical-2d-3d}
\BIBentryALTinterwordspacing
S.~Shan, X.~Guo, W.~Yan, E.~I. Chang, Y.~Fan, and Y.~Xu, ``Unsupervised
  end-to-end learning for deformable medical image registration,'' \emph{CoRR},
  vol. abs/1711.08608, 2017. [Online]. Available:
  \url{http://arxiv.org/abs/1711.08608}
\BIBentrySTDinterwordspacing

\bibitem{de2017end}
B.~D. de~Vos, F.~F. Berendsen, M.~A. Viergever, M.~Staring, and I.~I{\v{s}}gum,
  ``End-to-end unsupervised deformable image registration with a convolutional
  neural network,'' \emph{arXiv preprint arXiv:1704.06065}, 2017.

\bibitem{de2019deep}
B.~D. de~Vos, F.~F. Berendsen, M.~A. Viergever, H.~Sokooti, M.~Staring, and
  I.~I{\v{s}}gum, ``A deep learning framework for unsupervised affine and
  deformable image registration,'' \emph{Medical image analysis}, vol.~52, pp.
  128--143, 2019.

\bibitem{shen2019networks}
Z.~Shen, X.~Han, Z.~Xu, and M.~Niethammer, ``Networks for joint affine and
  non-parametric image registration,'' in \emph{Proceedings of the IEEE
  Conference on Computer Vision and Pattern Recognition}, 2019, pp. 4224--4233.

\bibitem{voxelmorph}
G.~Balakrishnan, A.~Zhao, M.~R. Sabuncu, J.~Guttag, and A.~V. Dalca, ``An
  unsupervised learning model for deformable medical image registration,'' in
  \emph{Proceedings of the IEEE conference on computer vision and pattern
  recognition}, 2018, pp. 9252--9260.

\bibitem{balakrishnan2019voxelmorph}
------, ``Voxelmorph: a learning framework for deformable medical image
  registration,'' \emph{IEEE transactions on medical imaging}, 2019.

\bibitem{krebs2019learning}
J.~Krebs, H.~e~Delingette, B.~Mailh{\'e}, N.~Ayache, and T.~Mansi, ``Learning a
  probabilistic model for diffeomorphic registration,'' \emph{IEEE transactions
  on medical imaging}, 2019.

\bibitem{krebs2018unsupervised}
J.~Krebs, T.~Mansi, B.~Mailh{\'e}, N.~Ayache, and H.~Delingette, ``Unsupervised
  probabilistic deformation modeling for robust diffeomorphic registration,''
  in \emph{Deep Learning in Medical Image Analysis and Multimodal Learning for
  Clinical Decision Support}.\hskip 1em plus 0.5em minus 0.4em\relax Springer,
  2018, pp. 101--109.

\bibitem{jaderberg2015spatial}
M.~Jaderberg, K.~Simonyan, A.~Zisserman \emph{et~al.}, ``Spatial transformer
  networks,'' in \emph{Advances in Neural Information Processing Systems},
  2015, pp. 2017--2025.

\bibitem{rudin1992nonlinear}
L.~I. Rudin, S.~Osher, and E.~Fatemi, ``Nonlinear total variation based noise
  removal algorithms,'' \emph{Physica D: nonlinear phenomena}, vol.~60, no.
  1-4, pp. 259--268, 1992.

\bibitem{ronneberger2015u}
O.~Ronneberger, P.~Fischer, and T.~Brox, ``U-net: Convolutional networks for
  biomedical image segmentation,'' in \emph{International Conference on Medical
  Image Computing and Computer-Assisted Intervention}.\hskip 1em plus 0.5em
  minus 0.4em\relax Springer, 2015, pp. 234--241.

\bibitem{godard2017unsupervised}
C.~Godard, O.~Mac~Aodha, and G.~J. Brostow, ``Unsupervised monocular depth
  estimation with left-right consistency,'' in \emph{CVPR}, vol.~2, no.~6,
  2017, p.~7.

\bibitem{rueckert1999nonrigid}
D.~Rueckert, L.~I. Sonoda, C.~Hayes, D.~L. Hill, M.~O. Leach, and D.~J. Hawkes,
  ``Nonrigid registration using free-form deformations: application to breast
  mr images,'' \emph{IEEE transactions on medical imaging}, vol.~18, no.~8, pp.
  712--721, 1999.

\bibitem{tensorflow}
\BIBentryALTinterwordspacing
M.~Abadi, A.~Agarwal, P.~Barham, E.~Brevdo, Z.~Chen, C.~Citro, G.~S. Corrado,
  A.~Davis, J.~Dean, M.~Devin, S.~Ghemawat, I.~Goodfellow, A.~Harp, G.~Irving,
  M.~Isard, Y.~Jia, R.~Jozefowicz, L.~Kaiser, M.~Kudlur, J.~Levenberg,
  D.~Man\'{e}, R.~Monga, S.~Moore, D.~Murray, C.~Olah, M.~Schuster, J.~Shlens,
  B.~Steiner, I.~Sutskever, K.~Talwar, P.~Tucker, V.~Vanhoucke, V.~Vasudevan,
  F.~Vi\'{e}gas, O.~Vinyals, P.~Warden, M.~Wattenberg, M.~Wicke, Y.~Yu, and
  X.~Zheng, ``{TensorFlow}: Large-scale machine learning on heterogeneous
  systems,'' 2015, software available from tensorflow.org. [Online]. Available:
  \url{https://www.tensorflow.org/}
\BIBentrySTDinterwordspacing

\bibitem{adam-optimizer}
\BIBentryALTinterwordspacing
D.~P. Kingma and J.~Ba, ``Adam: {A} method for stochastic optimization,''
  \emph{CoRR}, vol. abs/1412.6980, 2014. [Online]. Available:
  \url{http://arxiv.org/abs/1412.6980}
\BIBentrySTDinterwordspacing

\bibitem{lits-data}
\BIBentryALTinterwordspacing
``Liver tumor segmentation challenge,'' 2017. [Online]. Available:
  \url{http://lits-challenge.com/}
\BIBentrySTDinterwordspacing

\bibitem{miccai-data}
T.~Heimann, B.~van Ginneken, M.~A. Styner, Y.~Arzhaeva, V.~Aurich, C.~Bauer,
  A.~Beck, C.~Becker, R.~Beichel, G.~Bekes, F.~Bello, G.~Binnig, H.~Bischof,
  A.~Bornik, P.~M.~M. Cashman, Y.~Chi, A.~Cordova, B.~M. Dawant, M.~Fidrich,
  J.~D. Furst, D.~Furukawa, L.~Grenacher, J.~Hornegger, D.~KainmÜller, R.~I.
  Kitney, H.~Kobatake, H.~Lamecker, T.~Lange, J.~Lee, B.~Lennon, R.~Li, S.~Li,
  H.~P. Meinzer, G.~Nemeth, D.~S. Raicu, A.~M. Rau, E.~M. van Rikxoort,
  M.~Rousson, L.~Rusko, K.~A. Saddi, G.~Schmidt, D.~Seghers, A.~Shimizu,
  P.~Slagmolen, E.~Sorantin, G.~Soza, R.~Susomboon, J.~M. Waite, A.~Wimmer, and
  I.~Wolf, ``Comparison and evaluation of methods for liver segmentation from
  ct datasets,'' \emph{IEEE Transactions on Medical Imaging}, vol.~28, no.~8,
  pp. 1251--1265, Aug 2009.

\bibitem{loni}
D.~W. Shattuck, M.~Mirza, V.~Adisetiyo, C.~Hojatkashani, G.~Salamon, K.~L.
  Narr, R.~A. Poldrack, R.~M. Bilder, and A.~W. Toga, ``Construction of a 3d
  probabilistic atlas of human cortical structures,'' \emph{Neuroimage},
  vol.~39, no.~3, pp. 1064--1080, 2008.

\bibitem{adni}
S.~G. Mueller, M.~W. Weiner, L.~J. Thal, R.~C. Petersen, C.~R. Jack, W.~Jagust,
  J.~Q. Trojanowski, A.~W. Toga, and L.~Beckett, ``Ways toward an early
  diagnosis in alzheimer’s disease: the alzheimer’s disease neuroimaging
  initiative (adni),'' \emph{Alzheimer's \& Dementia}, vol.~1, no.~1, pp.
  55--66, 2005.

\bibitem{abide}
A.~Di~Martino, C.-G. Yan, Q.~Li, E.~Denio, F.~X. Castellanos, K.~Alaerts, J.~S.
  Anderson, M.~Assaf, S.~Y. Bookheimer, M.~Dapretto \emph{et~al.}, ``The autism
  brain imaging data exchange: towards a large-scale evaluation of the
  intrinsic brain architecture in autism,'' \emph{Molecular psychiatry},
  vol.~19, no.~6, p. 659, 2014.

\bibitem{adhd}
\BIBentryALTinterwordspacing
P.~Bellec, C.~Chu, F.~Chouinard-Decorte, Y.~Benhajali, D.~S. Margulies, and
  R.~C. Craddock, ``The neuro bureau adhd-200 preprocessed repository,''
  \emph{NeuroImage}, vol. 144, pp. 275 -- 286, 2017, data Sharing Part II.
  [Online]. Available:
  \url{http://www.sciencedirect.com/science/article/pii/S105381191630283X}
\BIBentrySTDinterwordspacing

\end{thebibliography}


\end{document}